\documentclass[11pt]{article}

\usepackage[final]{acl}

\usepackage{times}
\usepackage{latexsym}

\usepackage[T1]{fontenc}

\usepackage[utf8]{inputenc}

\usepackage{microtype}

\usepackage{inconsolata}
\usepackage{amsmath,amssymb}
\usepackage{graphicx}
\usepackage{adjustbox}
\usepackage{caption}
\usepackage{siunitx}
\usepackage{booktabs}
\usepackage{makecell}
\usepackage{graphicx}
\usepackage{subcaption}

\def\eqref#1{equation~\ref{#1}}

\def\1{\bm{1}}

\def\rvu{{\mathbf{i}}}

\def\rvu{{\mathbf{u}}}

\DeclareMathAlphabet{\mathsfit}{\encodingdefault}{\sfdefault}{m}{sl}
\SetMathAlphabet{\mathsfit}{bold}{\encodingdefault}{\sfdefault}{bx}{n}

\def\gD{{\mathcal{D}}}

\def\gH{{\mathcal{H}}}

\def\gN{{\mathcal{N}}}

\def\gV{{\mathcal{V}}}

\newcommand{\R}{\mathbb{R}}

\newcommand{\best}[1]{{\bfseries\tablenum{#1}}}
\usepackage{booktabs}   
\usepackage{multirow}   
\usepackage{makecell}   
\usepackage{array}
\usepackage{siunitx}

\usepackage{xcolor}
\newif\ifdraft\drafttrue

\title{From Heads to Neurons: Causal Attribution and Steering \\ in Multi-Task Vision–Language Models}

\author{
    Qidong Wang$^1$, Junjie Hu$^2$, Ming Jiang$^2$\thanks{corresponding author} \\
    $^1$Tongji University, $^2$University of Wisconsin-Madison\\
    \texttt{wang\_qidong@tongji.edu.cn}, \texttt{\{junjie.hu, ming.jiang\}@wisc.edu}
}

\begin{document}
\maketitle

\begin{abstract}
Recent work has increasingly explored neuron-level interpretation in vision-language models (VLMs) to identify neurons critical to final predictions. However, existing neuron analyses generally focus on single tasks, limiting the comparability of neuron importance across tasks. Moreover, ranking strategies tend to score neurons in isolation, overlooking how task-dependent information pathways shape the write-in effects of feed-forward network (FFN) neurons. This oversight can exacerbate neuron polysemanticity in multi-task settings, introducing noise into the identification and intervention of task-critical neurons. In this study, we propose HONES (\textbf{H}ead-\textbf{O}riented \textbf{N}euron \textbf{E}xplanation \& \textbf{S}teering), a gradient-free framework for task-aware neuron attribution and steering in multi-task VLMs. HONES ranks FFN neurons by their causal write-in contributions conditioned on task-relevant attention heads, and further modulates salient neurons via lightweight scaling. Experiments on four diverse multimodal tasks and two popular VLMs show that HONES outperforms existing methods in identifying task-critical neurons and improves model performance after steering. Our source code is released at: \url{https://github.com/petergit1/HONES}.
\end{abstract}

\section{Introduction}
\label{sec:introduction}

Large vision-language models (VLMs) have demonstrated strong multi-task capabilities across a wide range of vision-language applications, including visual question answering (VQA) \citep{liu2023visual,dai2023instructblip}, optical character recognition (OCR) \citep{ye-etal-2023-ureader,hu-etal-2025-mplug}, and image captioning \citep{li2023blip2,liu2025capability}. Despite their impressive performance, the internal decision-making process of these models remains opaque, as multiple capabilities are entangled within shared parameters, hindering error attribution and limiting reliability and controllability in real deployments. To address this gap, recent studies have increasingly focused on model interpretability, particularly neuron-level analysis, which offers fine-grained and actionable insights for model diagnosis and editing \citep{lin2025-mmfm-mi-survey,shu-etal-2025-survey,meng2022rome}.

Existing neuron analysis largely focuses on large language models (LLMs) \citep{zhao2023explainability,tang-etal-2024-language,yu-ananiadou-2024-neuron}. Recently, this line of work has begun to extend to multimodal settings \citep{huang2024miner,huo-etal-2024-mmneuron,pach2025sae_vlm}. Research in this area primarily follows two strands: (1) ranking neurons to identify those most salient to model prediction \citep{wang-etal-2025-attribution,dang2024mllm_survey}, and (2) analyzing the semantic information encoded by individual neurons \citep{pach2025sae_vlm,sajjad-etal-2022-neuron}. 

Despite remarkable progress, existing neuron analysis methods face two key limitations. First, they typically focus on interpreting neurons within a single task (e.g., VQA), leaving the comparability of these interpretability methods across tasks underexplored. This issue becomes particularly pronounced for tasks with distinct characteristics and heterogeneous outputs, such as question answering versus image–text matching. Second, many approaches analyze neurons in isolation, which leads to high computational cost, especially in large VLMs with expansive FFN layers, and becomes even more prohibitive in long-context multimodal settings. Moreover, treating neurons independently while overlooking task-relevant routing context from attention heads may exacerbate polysemanticity in multi-task VLMs \citep{oikarinen2024linear,haider2025ranges}. As a result, neurons participating in multiple routing paths can receive inflated importance scores, ultimately leading to noisier identification of truly task-relevant neurons.

To address the aforementioned issues, we propose \textbf{HONES} (Head-Oriented Neuron Explanation \& Steering), a unified, gradient-free framework for context-guided neuron attribution in multi-task VLMs, enabling consistent interpretability across heterogeneous readouts. Specifically, HONES builds upon a structured causal view of model computation, where attention heads select and route task-critical inputs, while downstream FFN neurons write the routed information into the residual stream for final prediction \citep{elhage2021mathematical}. Following this consideration, we first localize task-critical attention heads via causal interventions and then use these routing signals to guide downstream neuron attribution. With our findings, we further introduce a lightweight steering method that freezes the backbone and learns only sparse scaling factors over the identified task-critical neurons, enabling controlled task-specific improvements.

We perform HONES on four diverse multimodal tasks (i.e., VQA, OCR, captioning, and image-to-text retrieval) and evaluate it on two representative VLMs (i.e., LLaVA and Qwen). Our key findings include: (1) HONES outperforms state-of-the-art neuron ranking methods in identifying task-critical neurons across all tasks and both VLMs; (2) task-critical neurons present task-dependent layer preferences; (3) key neurons shared across multiple tasks are more salient than task-specific ones, particularly those overlapping with VQA; and (4) our lightweight steering method effectively improves performance across all four tasks on both VLMs.

Overall, our main contributions are as follows:
\begin{itemize}
\item We propose \textbf{HONES}, a head-conditioned and gradient-free framework for neuron-level causal attribution and steering in multi-task VLMs, where FFN neurons are identified under task-relevant routing context rather than scored in isolation.
  \item We uncover a variety of novel insights into task-critical neuron patterns that advance the understanding of VLM mechanisms in cross-task generalization. 
  \item We apply HONES to improve model performance via lightweight steering on the identified task-critical neurons. Experiments on four diverse multimodal tasks and two VLMs show consistent improvements.
\end{itemize}

\section{Related Work}
\label{sec:related}

\paragraph{Large VLMs.} Existing large VLMs typically combine a powerful visual encoder with an LLM. Popular training strategies include in-context multimodal conditioning like Flamingo \citep{alayrac2022flamingo} and visual instruction tuning for general-purpose task following, such as LLaVA \citep{liu2023visual} and InstructBLIP \citep{dai2023instructblip}. These models, especially open-source ones, have made steady progress in supporting higher-resolution visual inputs and document-centric understanding, as demonstrated by InternVL \citep{chen2024internvl}, mPLUG-Owl2 \citep{ye2023mplugowl2}, and Qwen2.5-VL \citep{bai2025qwen25vl}. Despite offering similar functionality, they differ in how visual signals are injected and how multi-task behaviors are routed internally, motivating mechanistic analysis across backbones. In this work, we focus on two representative models—LLaVA-1.5-7B \citep{liu2023visual} and Qwen2.5-VL-7B \citep{bai2025qwen25vl} to evaluate the generalizability of our findings and steering method across different model architectures.

\paragraph{Neuron-level Interpretability.} 

Existing neuron-level interpretability in VLMs largely inherits strategies from LLM-based studies. For example, prior work views FFN/MLP blocks as memory- and computation-like units that ``write'' into the residual stream, localize key neurons associated with facts/capabilities, and validate their causal roles through targeted interventions \citep{geva2021keyvalue,dai2022knowledgeneurons}. Recently, dictionary learning and sparse autoencoders (SAEs) have highlighted that \emph{features} (rather than individual neurons) offer stable analysis units, mitigating polysemanticity and enhancing interpretability \citep{bricken2023monosemantic}.

Building on this foundation, existing neuron-level studies for VLMs can be organized into two complementary streams: (1) causal analysis of model structure and (2) semantic analysis of latent representations. The former focuses on identifying which components causally drive the output behavior, and includes intervention-based analyses and activation-based neuron discovery followed by ablation or gating validation (e.g., domain-specific neurons~\citep{huo-etal-2024-mmneuron}, modality-specific neurons~\citep{huang2024miner}, and culture-sensitive neurons~\citep{zhao-etal-2025-culture}). It also covers readout-aligned neuron scoring, including prediction-probability-change methods~\citep{yu-ananiadou-2024-neuron}, as well as broader attribution methods that anchor neuron importance to output-side changes, yielding more direct and testable contribution mappings~\citep{schwettmann_2023_iccv,pan-etal-2024-finding,fang-etal-2024-nam,wang-etal-2025-attribution}. Relatedly, MultEdit \cite{basu2024understanding} provides complementary mechanistic evidence by editing early causal MLP blocks in multimodal LLMs. The latter is mainly represented by SAE-based sparse dictionary learning, which learns sparse and more monosemantic feature factors to characterize how high-level semantics are organized in representation space, and has been extended to visual or vision-language representations to support interpretability analysis and steering \citep{pach2025sae_vlm,lim2025patchsae}.

Our work differs from existing methods in three aspects. First, unlike intervention-based analyses that rely on token-level objectives and counterfactual constructions, HONES provides a unified scoring interface across heterogeneous task readouts, which is particularly useful for open-ended generation and retrieval ranking. Second, unlike readout-aligned attribution methods, HONES conditions neuron scoring on localized evidence-routing attention heads, yielding cleaner and more comparable neuron sets under shared parameters. Third, HONES is gradient-free and scalable: once key routing heads are identified, it requires only a constant number of additional forward passes and computes all FFN neuron scores jointly in a vectorized manner, avoiding the per-unit patching bottleneck of fine-grained causal methods. Compared with feature-level approaches such as SAEs, HONES is also model-native and directly supports causal attribution and lightweight steering on the original backbone without additional feature learning.

\section{Preliminaries}
\label{sec:prelim}

\paragraph{Multi-task VLM.}
We consider a multi-task VLM with parameters $\theta$ over heterogeneous tasks
$\mathcal{T}=\{T_{\mathrm{vqa}},T_{\mathrm{ocr}},T_{\mathrm{cap}},T_{\mathrm{ret}},\ldots\}$.
For a task $t\in\mathcal{T}$, each example is a labeled pair $(x,y)\in\mathcal{D}_t$,
where the input is a multimodal sequence $x=[x_v,x_t]$ (visual and textual tokens),
and $y$ is the task-specific ground-truth output.
Given $x$, the model produces a task prediction $\hat y$
(e.g., short answers for VQA, character sequences for OCR, free-form captions for captioning,
or decisions/rankings for retrieval). Note that our focus differs from multi-domain investigations, which typically study a fixed task across substantially different visual distributions (e.g., natural to medical imagery)~\citep{ huo-etal-2024-mmneuron}. In contrast, we focus on heterogeneous tasks within a shared visual domain, enabling controlled cross-task comparison of neuron attribution without confounding from domain shift.

\paragraph{Neuron Definition.}
Following~\citet{huo-etal-2024-mmneuron}, we focus on the two FFN layers at a Transformer layer $\ell$, denoted as $\mathbf{W}^{(\ell)}_\text{up}\in\mathbb{R}^{d\times d_\text{ff}}$ and $\mathbf{W}^{(\ell)}_\text{down}\in\mathbb{R}^{d_\text{ff}\times d}$. Let $\mathbf{z}^{(\ell)} \in \mathbb{R}^{d_\text{ff}}$ be the intermediate activations of the last token in $x$ after the first FFN at layer $\ell$, which directly affects next-token prediction in autoregressive decoding. We then define the $i$-th neuron at layer $\ell$ as the $i$-th element of $\mathbf{z}^{(\ell)}$, denoted as $z^{(\ell)}_{i} \in \mathbb{R}$, which is associated with its input weight $\mathbf{W}^{(\ell)}_{\mathrm{up}}[:,i]$ and output weight $\mathbf{W}^{(\ell)}_{\mathrm{down}}[i,:]$.
This definition ties a neuron to a concrete, manipulable computation in the model's FFN parameters.
Formally, we denote the index set of all neurons in $\theta$ as $\mathcal{U} = \{ (\ell, i) \mid \ell \in [1,L], i \in [1, d_{\mathrm{ff}}] \}$. 

\paragraph{Unified Dataset and Splits.}
We start from a unified multi-task dataset $\mathcal{D}$ constructed on a shared image set.
For each task $t$, we derive the task-specific labeled set $\mathcal{D}_t\subseteq \mathcal{D}$ by keeping the same image
$x_v$ and only changing the task instruction $x_t$ to form $x$ with the corresponding label $y$.
We split each $\mathcal{D}_t$ into three disjoint subsets:
$\mathcal{D}_t^{\mathrm{disc}}$, $\mathcal{D}_t^{\mathrm{dev}}$, and $\mathcal{D}_t^{\mathrm{test}}$,
with no overlap in images across splits.
We use $\mathcal{D}_t^{\mathrm{disc}}$ for head/neuron discovery (\S\ref{sec:head-local}, \S\ref{sec:neuron-attrib}), $\mathcal{D}_t^{\mathrm{dev}}$ for learning the scaling factors (\S\ref{sec:steering}),
and use $\mathcal{D}_t^{\mathrm{test}}$ only to \emph{verify} causal importance by masking the discovered neurons
and measuring the induced performance drop.

\paragraph{Instance-level Task Performance as a Scalar.}
We define an instance-level task performance scalar
$\mathcal{P}_t(x,y;\theta)\in\mathbb{R}$ where higher is better.
Concretely, $\mathcal{P}_t$ can be instantiated by the task-specific evaluation metric applied to the model prediction (e.g., VQAv2 accuracy, ANLS, BLEU-4, or NDCG@5 under a fixed evaluation protocol).
\begin{align}
\mathcal{P}_t(x,y;\theta) \;=\; \mathrm{Metric}_t\!\big(\hat y, y\big).
\end{align}

\begin{figure*}[t]
    \centering
    \includegraphics[width=\textwidth,height=0.18\textheight,keepaspectratio]{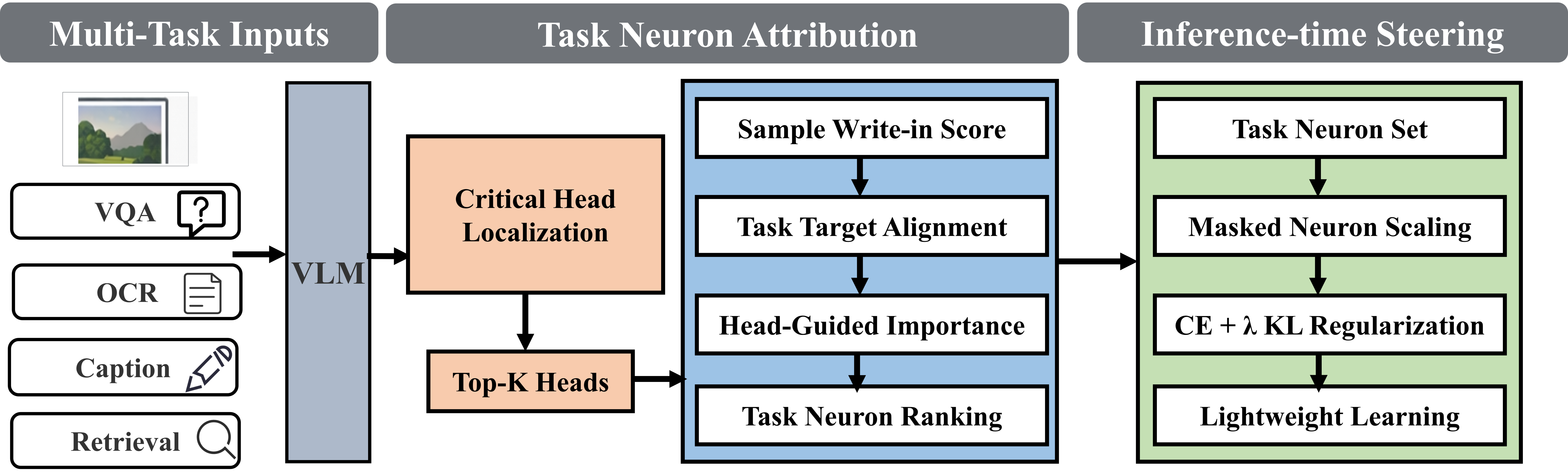}
    \caption{Overview of HONES. Left: Discovery task neurons via head-guided, readout-aligned write-in scoring. Right: Steering employs sparse activation scaling on a frozen backbone to enhance task performance.}
    \vspace{-4mm}
\label{fig:hones_framework}
\end{figure*}

\paragraph{Problem Definition}
For any neuron subset $\mathcal{S}\subset\mathcal{U}$, let $\mathcal{I}_{\mathcal{S}}$ be a masking operator
that suppresses activations of neurons in $\mathcal{S}$ at inference time, yielding an intervened model
$\theta_{\mathcal{I}_{\mathcal{S}}}$.
We quantify the causal importance of $\mathcal{S}$ on a labeled example $(x,y)$ by the performance drop:
\begin{align} \nonumber
\Delta \mathcal{P}_t\big((x,y);\mathcal{S}\big)
=
\mathcal{P}_t(x,y;\theta)
-
\mathcal{P}_t\!\big(x,y;\theta_{\mathcal{I}_{\mathcal{S}}}\big).
\end{align}

Our goal is to identify, for each task $t\in\mathcal{T}$, a compact set $\mathcal{N}_t^*$ of $K$ task-critical neurons whose causal intervention induces
a large expected performance drop under the task utility $\mathcal{P}_t$ for data in $\mathcal{D}_t^\text{test}$, i.e., masking these neurons should significantly reduce task performance.

We discover such neurons on the discovery split $\mathcal{D}_t^{\mathrm{disc}}$ and evaluate the induced performance drop $\Delta\mathcal{P}_t$
on the held-out test split $\mathcal{D}_t^{\mathrm{test}}$ to assess generalization.

\section{Method}
\label{sec:method}

\paragraph{Overview.}
Figure~\ref{fig:hones_framework} overviews \textsc{HONES}, a two-stage framework for readout-aligned neuron discovery
and causal steering in multi-task VLMs.
For \textbf{discovery}, we localize a sparse set of task-critical attention heads $\mathcal{H}_t^*$ via causal head interventions (\S\ref{sec:head-local}),
then identify task-critical FFN neurons by measuring each neuron's \emph{task-target-aligned} causal write-in contribution
conditioned on $\mathcal{H}_t^*$ using \textit{direct vocabulary projection} (\S\ref{sec:neuron-attrib}).
In \textbf{steering} (\S\ref{sec:steering}), we freeze the model $\theta$ and learn only sparse neuron-wise scaling factors on $\mathcal{N}_t^*$ with a KL-regularized objective, which both \emph{verifies} causality and enables controllable task improvements.

\subsection{Critical Head Localization}
\label{sec:head-local}

Our primary objective is to identify task-critical FFN neurons. 
Given that FFN write-ins are routed and aggregated via self-attention mechanisms, we first localize a sparse set of critical ``routing nodes'' (attention heads) prior to neuron-level attribution. 
This step is essential to constrain the search space and isolate effective computational pathways.

To this end, we adopt V-SEAM~\cite{wang-etal-2025-v}, a state-of-the-art causal head localization method with a mean-replacement intervention operator.
Let $\mathbf{o}^{(h)}(x)$ be the output of head $h$ in a certain layer.
The mean-replacement intervention $\mathcal{I}_{h}$ replaces this head's output with the mean of the remaining $H-1$ heads:
\begin{equation}
\label{eq:mean_replacement_head}
\begin{aligned}
\tilde{\mathbf{o}}^{(h)}(x)
&=
\frac{1}{H-1}\sum_{h'\neq h}\mathbf{o}^{(h')}(x).\\
\end{aligned}
\end{equation}
We then update the target head output by setting $\mathbf{o}^{(h)}(x)\leftarrow \tilde{\mathbf{o}}^{(h)}(x)$ during the forward pass.
Compared with hard zero-masking, mean-replacement reduces out-of-distribution artifacts.

\paragraph{Head Importance Score.}
A head's importance score is computed by the expected degradation in the task utility over the dataset $\mathcal{D}^{\text{disc}}_t$:
\begin{equation}
\label{eq:head_score}
S_t(h) 
=
\mathbb{E}_{(x,y)\sim\mathcal{D}_t^{\mathrm{disc}}}
\Big[
\Delta\mathcal{P}_t\big((x,y);\mathcal{I}_{h}\big)
\Big].
\end{equation}
Here $\Delta\mathcal{P}_t\big((x,y);\mathcal{I}_{h}\big)
=
\mathcal{P}_t(x,y;\theta)
-
\mathcal{P}_t\big(x,y;\theta_{\mathcal{I}_{h}}\big)$,
where $\theta_{\mathcal{I}_{h}}$ denotes the model under intervention in Eq.~(\ref{eq:mean_replacement_head}).
We select the Top-$K_h$ heads sorted by $S_t(h)$ as $\mathcal{H}_t^*$.

\subsection{Fine-grained Localization: Head-Guided Neuron Attribution}
\label{sec:neuron-attrib}

\paragraph{Motivation and Formulation.}
After localizing the critical routing nodes (the critical head set $\mathcal{H}_t^*$), we trace back and identify the FFN neurons that provide causally relevant information along these pathways.
A common heuristic ranks neurons by activation magnitudes \cite{huang2024miner,xu2025deciphering}, but highly activated neurons can be polysemantic, inhibitory, or write information that is subsequently filtered out by attention routing, resulting in many false positives and unstable task-critical sets.
Our intuition follows the \emph{key--value memory} view of Transformer FFNs~\citep{geva2021keyvalue}: once triggered, a neuron writes a content-bearing update into the residual stream via its down-projection direction.
Formally, given $\mathcal{H}_t^*$ and the neuron full set $\mathcal{U}$, we aim to produce a globally ranked list and select the task-critical neuron set $\mathcal{N}_t^*$.

Our key idea is to first compute a sample-level contribution score $c_{\ell, i}(x, y; \theta)$ of a neuron $(\ell, i)$ given a sample $(x,y) \in\gD_t^\text{disc}$, then apply head-guided interventions using $h\in\gH^*_t$ to quantify an overall importance score $I_{\ell, i}$ over all instances in $\gD_t^\text{disc}$, and lastly select the top K neurons by $I_{\ell, i}$ to form the task-critical neuron set $\gN_t^*$. We detail these three steps below.

\paragraph{Sample-level Write-in Contribution $c_{\ell,i}(x,y; \theta)$.}
The goal here is to quantify the contribution of neuron $(\ell, i)$ in the model $\theta$ to the prediction of $y$ given $x$. 
With the notations in \S\ref{sec:prelim}, we consider an FFN neuron $(\ell, i)\in\mathcal{U}$ and obtain its scalar activation $z^{(\ell)}_{i}$ by a forward pass of the last token in $x$ using the model $\theta$. By downprojection in the second FFN, we compute the neuron's output vector written into the subsequent residual stream as:
\begin{align}
    \Delta \mathbf{r}^{(\ell)}_i =z^{(\ell)}_i \mathbf{W}^{(\ell)}_{\mathrm{down}}[i,:] \in \R^{d}.
\end{align}

To quantify the write-in contribution of this residual increase $\Delta\mathbf{r}^{(\ell)}_i$ to the next-token prediction, we apply the \textit{direct vocabulary projection} (DVP), which reuses the LM's prediction head to project the residual stream to the vocabulary space $\gV$. Specifically, let $\mathbf{U}\in\R^{|\gV|\times d}$ be the unembedding matrix in the LM's prediction head, where $\mathbf{u}_v=\mathbf{U}[v,:]$ is the unembedding vector of a token $v\in\gV$. We then compute the neuron's contribution to affect the probability of predicting $v$ by:
\begin{align} \label{eq:c_v}
    c_{\ell, i}(x, v; \theta) = \langle \Delta \mathbf{r}^{(\ell)}_i, \mathbf{u}_v \rangle
\end{align}

Since the ground-truth target $y$ may contain a varying number of tokens, we need to aggregate its tokens to compute the unembedding vector $\rvu_y$ of $y$ before computing $c_{\ell, i}(x, y; \theta)$. To this end, we propose to compute $\rvu_y$ for two categories of $y$.

1. \textit{Fixed-set Targets} in tasks like VQA, OCR and retrieval where $y$ is restricted to a fixed set of possible outputs. In this case, we compute the target vector $\rvu_y$ as the normalized mean unembedding vectors for all the tokens in $y$, i.e., $\rvu_y = \frac{\sum_{v\in y} \mathbf{u}_v}
{\left\|\sum_{v\in y} \mathbf{u}_v\right\|_2}.$

2. \textit{Open-ended Targets} in tasks like captioning where $y$ can be any valid sentences. As not all tokens in $y$ are equally important to estimate $\rvu_y$, we estimate the inverse document frequency (IDF) $\alpha(v)$ of each token $v$ in $y$ using the instances in $\gD_t^\text{disc}$; see the IDF measure in Appendix \ref{sec:IDF}. We compute an IDF-weighted semantic center by:
\begin{equation}
\label{eq:utgt_cap}
\mathbf{u}_{y}
=
\frac{\sum_{v\in y} \alpha(v)\,\mathbf{u}_v}
{\left\|\sum_{v\in y} \alpha(v)\,\mathbf{u}_v\right\|_2}.
\end{equation}

Similar to Eq.~(\ref{eq:c_v}), we can compute the sample-level write-in contribution as: 
\begin{align}
    c_{\ell, i}(x, y; \theta) = \langle \Delta \mathbf{r}^{(\ell)}_i, \mathbf{u}_y \rangle.
\end{align}

\paragraph{Head-guided Importance Score $I_{\ell, i}$.} To condition attribution on the critical routing pathways,
for each critical head $h\in\mathcal{H}_t^*$ we apply a head intervention $\mathcal{I}_h$ (\S\ref{sec:head-local}) during inference and recompute the neuron's contribution $c_{\ell,i}(x,y; \theta_{\mathcal{I}_h})$ using the intervened model $\theta_{\mathcal{I}_h}$.
We measure the contribution drop due to head-guided intervention with respect to the original model $\theta$ as
\begin{equation} \nonumber
\Delta c^{(h)}_{\ell,i}(x,y)
= [
c_{\ell,i}(x,y; \theta)
-
c_{\ell,i}(x,y; \theta_{\mathcal{I}_h}) ]_{+},
\end{equation}
where $[u]_+=\max(u,0)$ ensures to focus on neurons whose task-specific contribution drops when critical head routes are disrupted. We then aggregate the contribution drop across heads and data by a head-importance--weighted average:
\begin{equation}
\label{eq:head_guided_importance}
\begin{aligned}
I_{\ell,i}
&=
\sum_{h\in\mathcal{H}_t^*}
w_h\;
\mathbb{E}_{(x,y)\sim \mathcal{D}_t^{\mathrm{disc}}}\!\big[\Delta c^{(h)}_{\ell,i}(x,y)\big],\\
w_h
&=
\frac{S_t(h)}{\sum_{h'\in\mathcal{H}_t^*} S_t(h')},
\qquad
\sum_{h\in\mathcal{H}_t^*} w_h = 1,
\end{aligned}
\end{equation}
where $S_t(h)$ is the head importance in Eq.~(\ref{eq:head_score}). 

\paragraph{Final Ranking.}
We compute $I_{\ell,i}$ in Eq.~(\ref{eq:head_guided_importance}) on the discovery split $\mathcal{D}_t^{\mathrm{disc}}$ for all neurons $(\ell,i)\in\mathcal{U}$,
rank them globally, and select the Top-$K$ neurons to form the task-critical set:
\begin{equation}
\mathcal{N}_t^*
=
\mathrm{TopK}\!\big(I_{\ell,i}\;:\;(\ell,i)\in\mathcal{U}\big).
\end{equation}

\subsection{Inference-time Steering}
\label{sec:steering}

\paragraph{Neuron Mask and Scaling.} To \emph{verify} whether the neurons in $\gN_t^*$ are causally responsible for task $t$, we learn to steer all the detected neurons in $\gN_t^*$ using the validation split $\gD_t^{\mathrm{dev}}$, while keeping all backbone weights $\theta$ frozen. Specifically, we introduce a set of learnable scaling factors $\lambda_t = \{\lambda_{\ell,i} \mid (\ell,i)\in\gN_t^*\}$ and steer the model by scaling only the detected task-critical neurons, i.e., $\tilde{z}^{(\ell)}_i = z^{(\ell)}_i \times \lambda_{\ell,i}, \forall (\ell,i)\in\gN_t^*$. We denote the resulting neuron-scaled model for task $t$ as $\theta_{\lambda_t}$.

\paragraph{Task-specific Lightweight Learning.}
We learn $\lambda_t$ \emph{separately for each task} on $\mathcal{D}^{\mathrm{dev}}_t$,
while keeping all backbone weights in $\theta$ frozen.
Let $p_{\theta}(\cdot|x)$ and $p_{\theta_{\lambda_t}}(\cdot|x)$ be the next-token distributions computed using the original model and the neuron-scaled model.
We seek the optimal $\lambda_t$ by minimizing the task-specific loss and the KL term that regularizes the scaled model to stay close to the original model.
\begin{align}
\label{eq:steer_obj}
\min_{\lambda_t}
\;
\mathbb{E}_{(x,y)\in \mathcal{D}^{\mathrm{dev}}_t}
\Big[
&\mathcal{L}_{t}\big(x,y;\theta_{\lambda_t}\big) \\ \nonumber
&+\beta\,\mathrm{KL}\!\Big(
p_{\theta}(\cdot|x)\,\|\,p_{\theta_{\lambda_t}}(\cdot|x)
\Big)
\Big].
\end{align}

\begin{table*}[t]
\centering
\footnotesize
\renewcommand{\arraystretch}{1.12}
\setlength{\tabcolsep}{3.5pt}
\begin{tabular}{@{}lccccc ccccc@{}}
\toprule
\multirow{2}{*}{Strategy} &
\multicolumn{5}{c}{LLaVA-1.5-7B} &
\multicolumn{5}{c}{Qwen2.5-VL-7B} \\
\cmidrule(lr){2-6}\cmidrule(lr){7-11}
& VQA & OCR & Caption & Retrieval & Avg
& VQA & OCR & Caption & Retrieval & Avg \\
\midrule
AP~\mbox{\citep{huang2024miner}}       & 11.33 & 10.40 &  8.65 & 0.50 &  7.72  &  5.00 & 10.40 & 14.51 & 0.19 &  7.53 \\
MA~\mbox{\citep{xu2025deciphering}}    &  6.82 & 15.50 & 11.90 & 1.35 &  8.89  & 22.05 & 15.50 &  9.86 & 0.65 & 12.02 \\
APE~\mbox{\citep{huo-etal-2024-mmneuron}} &  3.20 & -1.87 & 12.20 & 0.90 &  3.61  & -2.36 & -1.87 &  6.20 & 0.08 &  0.51 \\
\midrule
HONES-NoHead     & 10.20 & \multicolumn{1}{c}{4.50} & \multicolumn{1}{c}{6.20} & \multicolumn{1}{c}{0.40} & \multicolumn{1}{c}{5.33}
                 & \multicolumn{1}{c}{11.00} & \multicolumn{1}{c}{8.75} & \multicolumn{1}{c}{10.50} & \multicolumn{1}{c}{2.60} & \multicolumn{1}{c}{8.21} \\
HONES-RandHead   &  9.70 &  6.30 & 10.00 & 1.50 &  6.88  & 13.00 & 11.50 & 18.00 & 3.75 & 11.56 \\
HONES-Gaussian   &  4.50 &  3.60 &  9.45 & 0.95 &  4.63  &  5.50 &  9.00 & 12.50 & 0.80 &  6.95 \\
RandNeuron       &  0.85 &  1.62 &  2.30 & 0.06 &  1.21  &  2.45 &  1.98 &  3.30 & 0.15 &  1.97 \\
\midrule
\textbf{HONES (Ours)} &
\textbf{27.30} & \textbf{19.00} & \textbf{19.80} & \textbf{7.43} & \textbf{18.38} &
\textbf{36.50} & \textbf{21.00} & \textbf{24.80} & \textbf{5.35} & \textbf{21.91} \\
\bottomrule
\end{tabular}
\caption{Relative performance drop (\%) after masking the top-1\% neurons selected by each ranking method.
Higher values indicate greater causal importance (negative values indicate improvement).}
\label{tab:neuron_main_two_models}
\vspace{-3mm}
\end{table*}

\begin{table}[t]
\centering
\normalsize
\renewcommand{\arraystretch}{1.10}
\setlength{\tabcolsep}{3pt}
\resizebox{\columnwidth}{!}{%
\begin{tabular}{lcc}
\toprule
\textbf{Method} & \textbf{LLaVA-1.5-7B} & \textbf{Qwen2.5-VL-7B} \\
\midrule
DLA \cite{elhage2021mathematical} & 7.40 & 9.80 \\
Group Patching \cite{zhang2023towards} & 18.00 & 20.05 \\
QRNCA \cite{chen2025identifying} & 20.80 & 24.50 \\
LLM-Knowledge \cite{yu-ananiadou-2024-neuron} & 16.50 & 19.40 \\
\midrule
\textbf{HONES (Ours)} & \textbf{27.30} & \textbf{36.50} \\
\bottomrule
\end{tabular}%
}
\caption{Comparison of neuron localization baselines on VQA, measured by the relative performance drop (\%) after masking the top-1\% neurons identified by each method. Larger drops indicate that the selected neurons are more causally important for task performance.}
\label{tab:vqa_extra_baselines}
\end{table}

\section{Experiments}
\label{sec:vlm-interpretations}

\subsection{Experimental Setting}
\label{sec:exp-setting}
\paragraph{Tasks and Data.} We focus on four popular vision-language tasks: VQA, OCR, image captioning (Caption), and image-to-text retrieval (Retrieval). To ensure comparability across tasks, we construct a unified multi-task benchmark based on the \texttt{train2014} split of \textbf{MS COCO}~\citep{lin2014coco}. Specifically, we identify the shared images from \textbf{COCO-Text}~\citep{veit2016cocotext}, \textbf{MSCOCO Captions}~\citep{chen2015cococaptions}, and \textbf{VQAv2}~\citep{goyal2017vqav2}, and curate a subset of 12,000 images, each of which contains human annotations for all tasks from the corresponding source benchmarks. Following the standard practice of prior work \cite{huo-etal-2024-mmneuron}, we split the data into 7K for neuron analysis, 2K for model-steering parameter tuning, and 3K for testing (see details in Appendix~\ref{app:dataset-details}).

\paragraph{Base VLMs and Baselines.} We employ two widely used VLMs as our base models: \textbf{LLaVA-1.5-7B}~\citep{liu2023visual} and \textbf{Qwen2.5-VL-7B}~\citep{bai2025qwen25vl}. To evaluate the effectiveness of HONES in identifying key neurons, we compare our method against three state-of-the-art activation-based neuron ranking methods across all four tasks: (1) activation probability \textbf{(AP)}~\citep{huang2024miner}, (2) mean activation \textbf{(MA)}~\citep{xu2025deciphering}, and (3) activation probability entropy \textbf{(APE)}~\citep{huo-etal-2024-mmneuron}. These methods are naturally comparable across tasks because they rely only on activation statistics and do not require task-specific supervision signals or readout-specific surrogate objectives. In addition to activation-statistic rankings, we also compare HONES against other popularly-used neuron ranking strategies, ranging from logit attribution to causal tracing to gradient-based attribution. We conduct these comparisons exclusively on VQA, as it provides naturally defined token-level supervision. Extending the same setup to tasks such as OCR, captioning, and retrieval would require task-specific aggregation or surrogate objectives, thereby reducing cross-task comparability. Our additional baselines include (4) direct logit attribution \textbf{(DLA)} ~\cite{elhage2021mathematical}, (5) activation patching with neuron groups \textbf{(Group Patching)}~\citep{zhang2023towards}, (6) \textbf{QRNCA}~\citep{chen2025identifying}, a gradient-based neuron ranking method, and (7) neuron-level knowledge attribution \textbf{(LLM-Knowledge)}~\citep{yu-ananiadou-2024-neuron}. To further assess the benefits of our neuron-level intervention strategy, we consider four control variants: random neuron masking \textbf{(RandNeuron)}, an ablation variant without head conditioning \textbf{(HONES--NoHead)}, HONES with random attention head masking \textbf{(HONES-RandHead)}, and HONES with Gaussian noise injection \textbf{(HONES-Gaussian)}.

Regarding model steering, we assess effectiveness and OOD generalizability by comparing HONES against matched-budget baselines:  (1) fixed uniform two-fold amplification on HONES-identified neurons \textbf{(Fixed-Amp)}, (2) grid-searched and uniform amplification on these neurons \textbf{(Grid-Search)}, (3) learnable scaling on random neurons \textbf{(RandNeuron)}, and (4) HONES steering without KL regularization \textbf{(HONES w/o KL)}. We also tested LoRA but found it ineffective under our low-budget setting.

\vspace{-1.5mm}\paragraph {Metrics.} We measure model performance on each task using standard metrics: accuracy for VQA, average normalized levenshtein similarity (ANLS) for OCR ~\citep{Biten_2019_ICCV}, BLEU-4 for Caption~\citep{papineni2002bleu}, and NDCG@5 for Retrieval~\citep{jarvelin2002cumulated}. Additional implementation details and prompt templates are provided in Appendix~\ref{app:dataset-details}.

\begin{figure}[t]
    \centering
    \includegraphics[width=0.95\columnwidth]{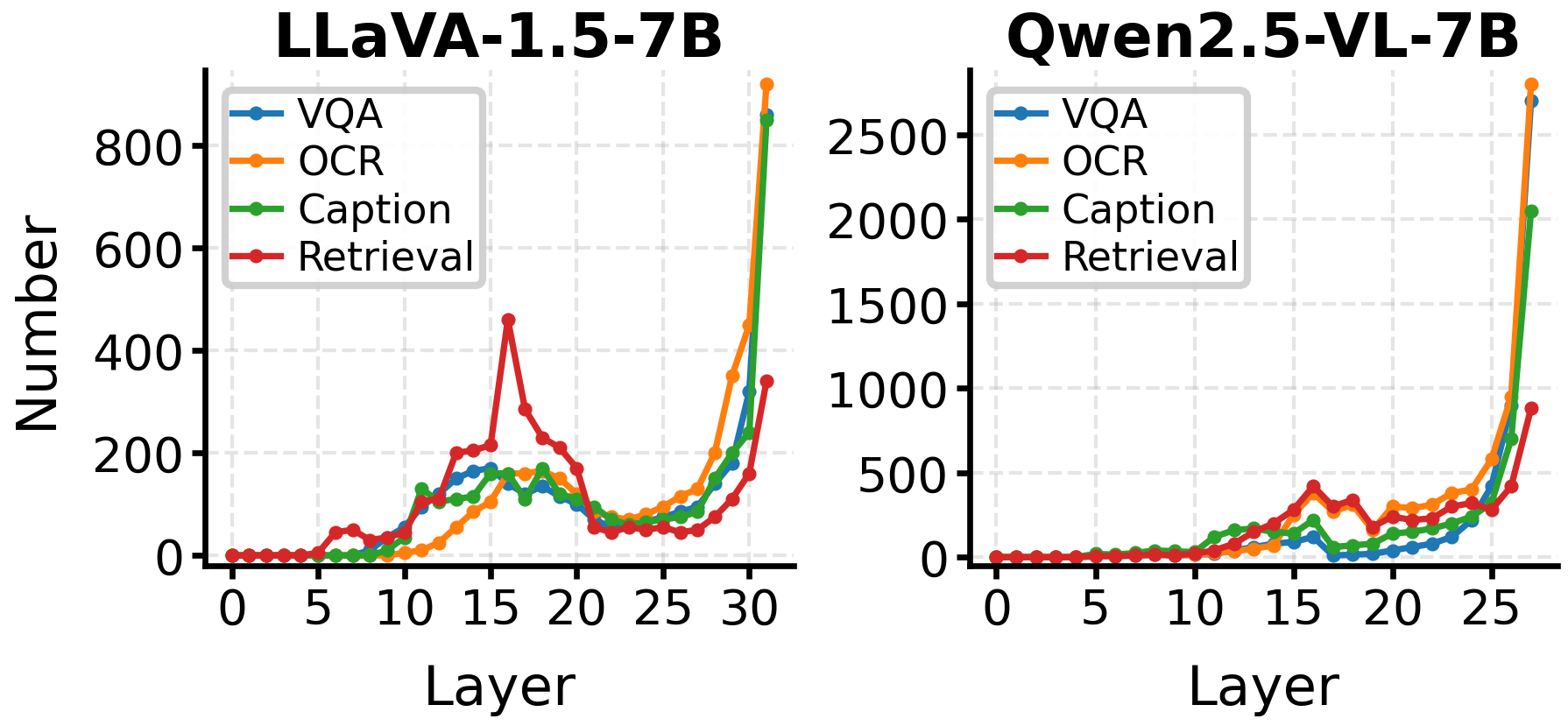}
 \caption{Layer-wise distribution of the top-1\% task-critical neurons across four tasks (VQA/OCR/Caption/Retrieval) for both models.}
    \label{fig:layer_neuron_dist}
    \vspace{-4mm}
\end{figure}

\subsection{Key Findings in Neuron Analysis}
\label{sec:key-findings}

\paragraph{Superior localization of key neurons with HONES compared to activation-based methods.}

Table~\ref{tab:neuron_main_two_models} and Table~\ref{tab:vqa_extra_baselines} show the relative performance drop induced by masking top-1\% neurons selected by each interpretation method across four tasks on two base VLMs. Based on empirical validation (see Appendix~\ref{sec:head-budget-ablation}), HONES identifies key neurons by ranking them using the top 30 attention heads in LLaVA and the top 25 attention heads in Qwen, respectively. The results show that our method consistently outperforms all baselines in identifying critical neurons on both base models, inducing an average performance drop of 18.38\% on LLaVA and 21.91\% on Qwen. Notably, masking neurons ranked by APE leads to performance improvements on the VQA and OCR tasks, suggesting that entropy-based neuron ranking may capture task-irrelevant or interfering neurons. Further comparison of neuron intervention strategies shows that masking top-ranked attention heads is substantially more effective for identifying critical neurons across tasks—yielding over a 10\% greater performance drop—than masking randomly ranked heads or applying Gaussian noise injection. Our results indicate that neuron importance is better captured by causally aligned information flow to the model’s readout than by raw activation magnitude, which is often confounded by polysemantic activity. We also assess HONES' performance on a larger backbone, \textbf{LLaVA-1.5-13B}, under the same four-task setting. The available results show trends consistent with those of the 7B models, with HONES remaining the strongest localization method. Full results are provided in Appendix~\ref{app:llava13b}. Beyond effectiveness, we further assess HONES in terms of localization efficiency on VQA against two representative baselines that are most relevant to our method, namely the gradient-based attribution method QRNCA and the intervention-based Group Patching. Our results show that HONES, as a gradient-free method, localizes key neurons substantially faster than its two counterparts (see Appendix~\ref{sec:appendix_efficiency}).

\begin{figure}[t]
    \centering
    \includegraphics[width=\columnwidth]{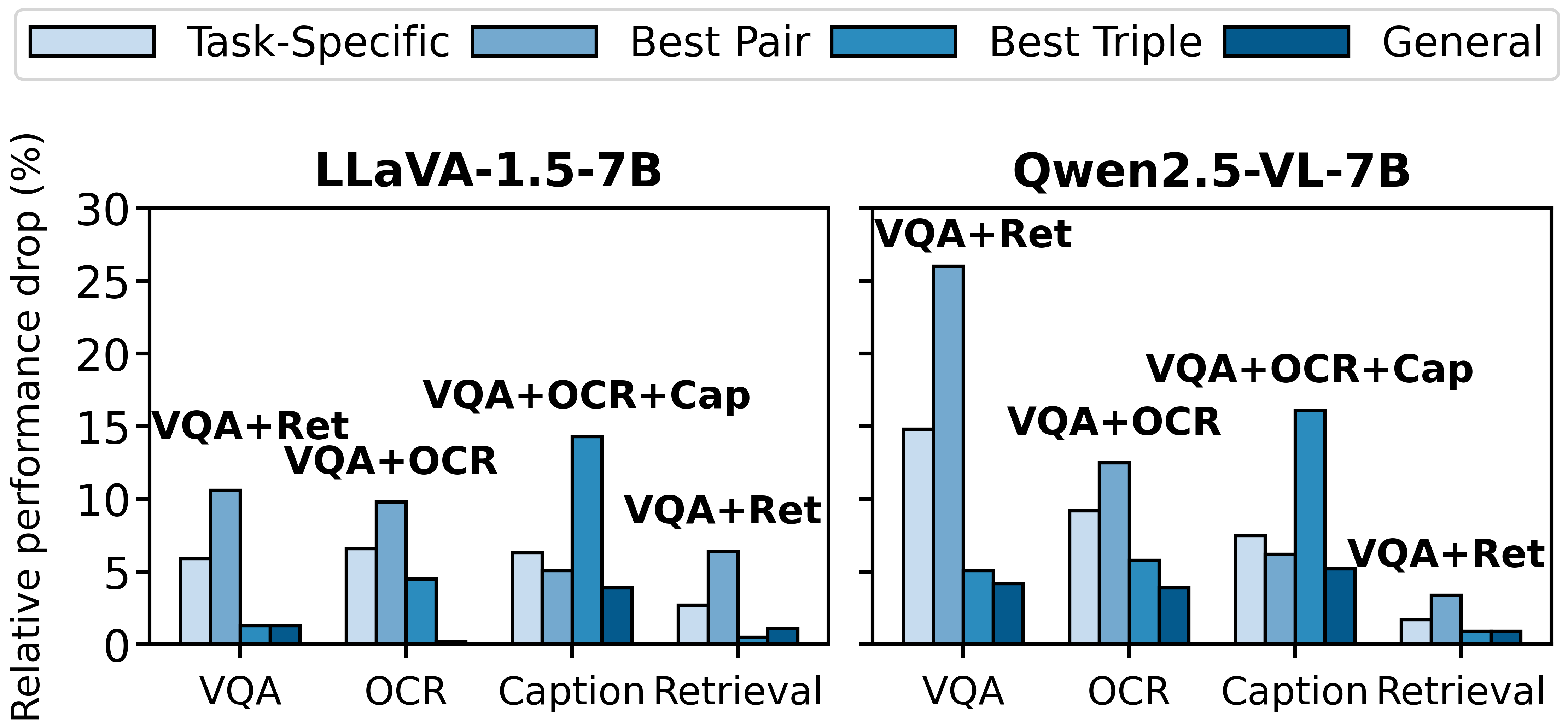}
   \caption{Cross-task neuron-group ablation results.
Bars denote the relative performance drop (\%) on each target task (x-axis) when masking the corresponding neuron group (legend): \emph{task-specific}, the most damaging two-task overlap (\emph{Best Pair}), the most damaging three-task overlap (\emph{Best Triple}), and the four-task shared \emph{General} group.
Text labels indicate the identities of \emph{Best Pair}/\emph{Best Triple} for each target task.}
\label{fig:cross_task_ablation_bars}
\vspace{-4mm}
\end{figure}

\begin{figure*}[t]
    \centering
    \includegraphics[width=\textwidth]{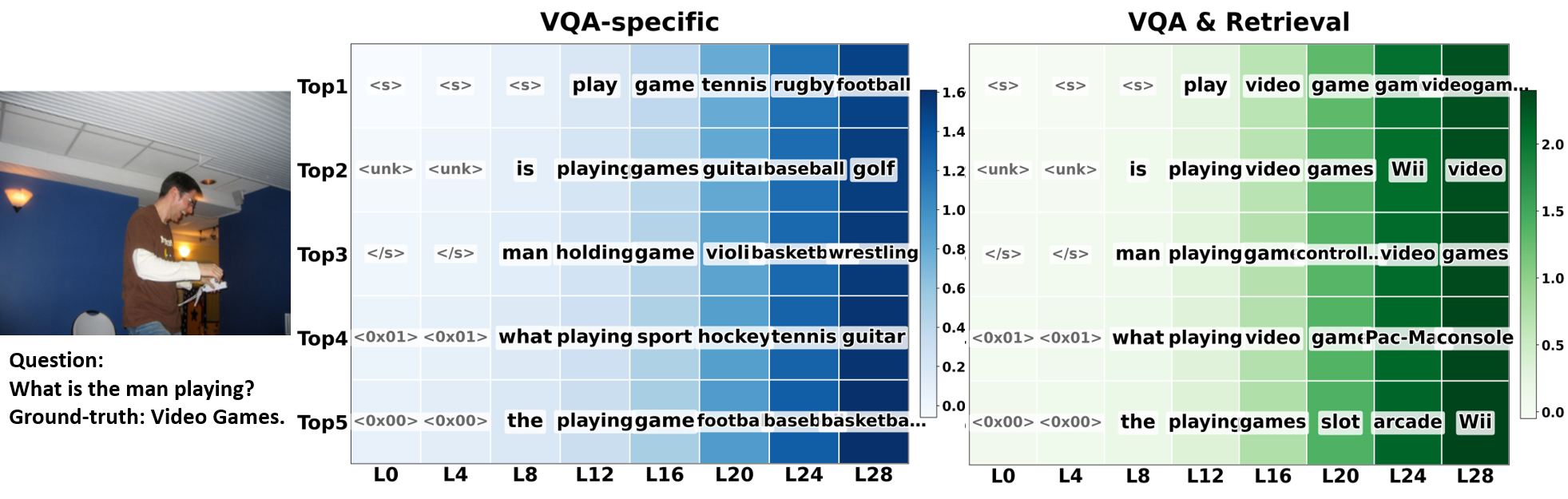}
    
\caption{VQA Logit Lens case study in LLaVA-1.5.
Rows show Top-5 tokens and columns are sampled every 4 layers; color indicates $\Delta$logit (baseline$-$masked) for the task-specific group and the VQA\&Retrieval shared group.}
    \label{fig:logitlens_vqa}
    \vspace{-4mm}
\end{figure*}

\paragraph{Key neurons exhibit task-dependent layer preferences.}  Inspired that VLMs perform different stages of multimodal information processing across layers, we further explore the layer-wise distribution of our identified top-1\% neurons per task, aiming to examine the processing stages at which these neurons emerge and their relationship to task characteristics. Specifically, we aggregate the number of identified neurons within each layer. This layer-wise count allows us to pinpoint the specific network depths where task-critical computations are most concentrated. Fig.~\ref{fig:layer_neuron_dist} displays the results for each base model. Overall, we observe that key neurons are predominantly concentrated in the middle layers (10–20) and deeper layers ($>$25). In contrast to the retrieval task, which tends to exhibit a peak in the middle layers, the other three tasks exhibit higher concentrations in deeper layers for both VLMs. Given that middle layers are generally associated with visual–text alignment, whereas deeper layers are more involved in answer decoding \cite{yu2025mllmstages}, these results suggest that retrieval relies more heavily on visual–text alignment, while text-based answer generation plays a larger role in the remaining three generation-oriented tasks.

\begin{table}[!t]
\centering
\footnotesize
\renewcommand{\arraystretch}{1.1}
\setlength{\tabcolsep}{3.5pt}
\resizebox{\columnwidth}{!}{%
\begin{tabular}{llccccc}
\toprule
\textbf{Model} & \textbf{Method} & \textbf{VQA} & \textbf{OCR} & \textbf{Caption} & \textbf{Retrieval} & \textbf{Avg} \\
\midrule
\multirow{6}{*}{LLaVA-1.5}
& Base              & 0.6520 & 0.5760 & 0.1285 & 0.9328 & 0.5723 \\
\cmidrule{2-7} 
& Fixed-Amp & 0.6610 & 0.5820 & 0.1290 & 0.9370 & 0.5773 \\
& Grid-Search       & 0.6660 & 0.5944 & 0.1315 & 0.9560 & 0.5870 \\
& RandNeuron       & 0.6530 & 0.5735 & 0.1260 & 0.9345 & 0.5718 \\
\cmidrule{2-7} 
& Ours w/o KL       & 0.6580 & 0.5790 & 0.1385 & 0.9365 & 0.5780 \\
& \textbf{Ours}     & \textbf{0.6733} & \textbf{0.6019} & \textbf{0.1407} & \textbf{0.9626} & \textbf{0.5946} \\
\midrule
\multirow{6}{*}{Qwen2.5-VL}
& Base              & 0.6750 & 0.6580 & 0.2240 & 0.9464 & 0.6259 \\
\cmidrule{2-7} 
& Fixed-Amp & 0.6773 & 0.6620 & 0.2280 & 0.9492 & 0.6291 \\
& Grid-Search       & 0.6803 & 0.6650 & 0.2290 & 0.9500 & 0.6311 \\
& RandNeuron       & 0.6760 & 0.6560 & 0.2210 & 0.9480 & 0.6253 \\
\cmidrule{2-7} 
& Ours w/o KL       & 0.6790 & 0.6610 & 0.2250 & 0.9475 & 0.6281 \\
& \textbf{Ours}     & \textbf{0.6907} & \textbf{0.6790} & \textbf{0.2335} & \textbf{0.9580} & \textbf{0.6403} \\
\bottomrule
\end{tabular}%
}
\caption{Effectiveness of neuron steering.
Fixed-Amp and Grid-Search are train-free amplifications; RandNeuron learns scaling on a random neuron set; Ours w/o KL removes KL regularization; \textbf{Ours} learns sparse scaling with KL.
Metrics: Acc.\ (VQA), ANLS (OCR), BLEU-4 (Caption), NDCG@5 (Retrieval).}
\label{tab:steer_effectiveness}
\vspace{-5mm}
\end{table}

\begin{table*}[t]
\centering
\small
\renewcommand{\arraystretch}{1.1}
\setlength{\tabcolsep}{4pt}

\begin{adjustbox}{max width=\textwidth}
\begin{tabular}{llccccc}
\toprule
\textbf{Model} & \textbf{Method} &
\textbf{GQA} & \textbf{TextVQA} & \textbf{Flickr$_{\text{Cap}}$} & \textbf{Flickr$_{\text{Ret}}$} & \textbf{Avg} \\
\midrule
\multirow{4}{*}{LLaVA-1.5}
& Base
& 0.610$\pm$0.001 & 0.470$\pm$0.001 & 0.097$\pm$0.001 & 0.863$\pm$0.002 & 0.510$\pm$0.001 \\
& Ours (zero-shot)
& 0.616$\pm$0.001 & 0.474$\pm$0.002 & 0.099$\pm$0.002 & 0.870$\pm$0.001 & 0.515$\pm$0.001 \\
& RandNeuron
& 0.610$\pm$0.001 & 0.469$\pm$0.002 & 0.097$\pm$0.001 & 0.864$\pm$0.002 & 0.510$\pm$0.001 \\
& \textbf{Ours (tuned)}
& \textbf{0.628$\pm$0.002} & \textbf{0.488$\pm$0.003} & \textbf{0.106$\pm$0.002} & \textbf{0.883$\pm$0.002} & \textbf{0.526$\pm$0.001} \\
\midrule
\multirow{4}{*}{Qwen2.5-VL}
& Base
& 0.632$\pm$0.001 & 0.596$\pm$0.002 & 0.163$\pm$0.002 & 0.888$\pm$0.002 & 0.570$\pm$0.001 \\
& Ours (zero-shot)
& 0.636$\pm$0.002 & 0.601$\pm$0.001 & 0.165$\pm$0.002 & 0.892$\pm$0.002 & 0.573$\pm$0.001 \\
& RandNeuron
& 0.632$\pm$0.001 & 0.596$\pm$0.002 & 0.163$\pm$0.001 & 0.889$\pm$0.001 & 0.570$\pm$0.001 \\
& \textbf{Ours (tuned)}
& \textbf{0.641$\pm$0.002} & \textbf{0.607$\pm$0.002} & \textbf{0.169$\pm$0.002} & \textbf{0.898$\pm$0.002} & \textbf{0.579$\pm$0.001} \\
\bottomrule
\end{tabular}
\end{adjustbox}

\vspace{-1mm}
\caption{OOD steering results (Accuracy on GQA, ANLS on TextVQA, BLEU-4 on FlickrCap, and NDCG@5 on FlickrRet).
\textbf{Ours (tuned)} tunes scalars on the identified task-critical neurons; \textit{RandNeuron} tunes scalars on randomly selected neurons with a matched budget; \textit{Ours (zero-shot)} directly transfers in-domain scales.}
\label{tab:ood_steering_placeholder}
\vspace{-6mm}
\end{table*}

\paragraph{VQA-shared neurons dominate cross-task saliency.} Examining key neurons across tasks, we find that a substantial number of these neurons are shared among multiple tasks. This motivates us to explore how the influence of these multi-task neurons compares to task-specific neurons across tasks, providing insight into the mechanisms underlying cross-task generalization and model sparsity. 

Given the top-1\% neurons identified by HONES per task, we group neurons based on the tasks across which they are shared. For each target task, we then calculate the relative performance drop resulting from masking each neuron group. Fig.~\ref{fig:cross_task_ablation_bars} presents the results for both base models. To conserve space, we only show the shared group that induces the largest drop when neurons are shared by two and three tasks, respectively (see full comparisons in Appendix ~\ref{sec:appendix_finding3}). Interestingly, multi-task shared neurons exhibit higher saliency than task-specific ones, particularly those overlapping with VQA, suggesting a hub effect whereby VQA-related neurons support a broad range of vision–language tasks. Pairwise shared neurons demonstrate dominant causal saliency across tasks, with the exception of captioning. We hypothesize that this deviation arises because captioning relies on holistic image understanding, while textual cues present in our curated images may bias the most salient neuron group toward OCR-related neurons. To verify our hypothesis, we repeat the analysis on the same images after removing textual cues. Following Appendix~\ref{sec:appendix_caption_text_control},  the most salient neuron group for the captioning task shifts back to the VQA–Caption pair.  We further examine patterns of key-neuron sharing across tasks on out-of-distribution datasets and observe the same qualitative trend, suggesting that the identified routing-and-neuron structures generalize across diverse data distributions (Appendix~\ref{app:ood_shared_groups}).

Furthermore, token-level Logit Lens analysis~\cite{neo2024towards} provides semantic corroboration. Specifically, we quantify the causal effect of a neuron group by the logit drop on target tokens, $\Delta \text{logit} = \text{logit}_{\text{base}} - \text{logit}_{\text{masked}}$. We compute this layer-wise by projecting the MLP layer outputs to the vocabulary space. By comparing the projected logits before and after masking the target neuron group, we inspect which candidate tokens lose the most support under intervention. As illustrated in Fig.~\ref{fig:logitlens_vqa}, VQA-specific neurons mainly strengthen coarse answer priors (e.g., generic category-level candidates), whereas the VQA\&Retrieval shared group selectively amplifies fine-grained, visually grounded tokens, effectively shifting predictions toward the ground truth. We observe consistent trends in OCR, captioning, and retrieval (Appendix~\ref{sec:appendix_logitlens}).

\subsection{Results of Neuron-level Model Steering}
\label{sec:model-editing}

\paragraph{Effectiveness.}

Table~\ref{tab:steer_effectiveness} displays the performance of HONES in neuron-level model steering compared against a variety of baselines. To improve steering efficiency, we restrict steering to the dominant salient neuron group per task identified in Sec.~\ref{sec:key-findings}, rather than intervening on the entire set of top-1\% neurons. Overall, HONES consistently outperforms all baselines across tasks on both backbone models. The gap to \textbf{RandNeuron} provides interventional evidence that task capability is concentrated in a small subset of neurons, while the gain over \textbf{Grid-Search} indicates that using neuron-wise modulation, rather than a single uniform amplifier, is necessary to fully exploit these causal pathways in practice.

\paragraph{OOD Generalization.}
To test the generalizability of our neuron steering method, we further evaluate the steered models on three out-of-distribution benchmarks for four tasks: GQA (VQA), TextVQA (OCR), and Flickr30k (Caption \& Retrieval). Similarly, we focus here exclusively on neurons belonging to the dominant salient group for each task.
As shown in Table~\ref{tab:ood_steering_placeholder}, directly transferring the in-domain learned scaling factors (\textit{Ours (zero-shot)}) already yields consistent gains.
Moreover, without re-localizing attention heads or re-identifying neurons, and using only 20\% OOD samples to learn scaling with evaluation on the remaining 80\% (averaged over 10 random splits; mean$\pm$std), our method consistently outperforms the random-neuron control.
This indicates that task capabilities are indeed concentrated and anchored in those sparse neurons from the dominant salient neuron group. Additional settings and statistical details are provided in Appendix ~\ref{sec:appendix_significance}.

\section{Conclusion}
\label{sec:conclusion}

We present \textbf{HONES}, a head-oriented, readout-aligned causal interpretability framework for multi-task VLMs.
By localizing task-critical routing heads and ranking MLP neurons via Causal Write-in Effect, HONES provides a unified and verifiable notion of neuron importance across heterogeneous task readouts.
Through a systematic analysis of LLaVA-1.5-7B and Qwen2.5-VL-7B, we find that activation statistics are often decoupled from causal criticality, task-critical computation follows a stable rise--valley--surge depth profile with task-dependent anchoring, and cross-task behaviors are mediated by a sparse VQA-centric \emph{Hub-and-Bridge} sharing structure. Our learning-based steering further operationalizes these discoveries by learning only a small set of neuron-wise scaling factors, enabling effective and stable control of task behaviors while keeping all backbone weights frozen.

\section*{Limitations}
\label{sec:limitations}

Despite providing new insights into interpretability and controllable interventions for large VLMs, our study has several limitations.

\begin{enumerate}
    \item \textbf{Model scale and architectural coverage.}
    To enable exhaustive layer-wise and neuron-level scanning, we focus on 7B dense backbones (LLaVA-1.5-7B and Qwen2.5-VL-7B).
    While both exhibit a consistent \emph{VQA-centric hub} pattern, it remains unclear whether this structure generalizes to substantially larger models (e.g., 70B+) or mixture-of-experts (MoE) architectures.
    As capacity increases, functional modularization may emerge differently or routing may become more distributed, which we leave for future work with more substantial compute budgets.

    \item \textbf{Task definition granularity.}
    For a unified cross-task analysis, we study four coarse task categories (VQA, OCR, Caption, Retrieval).
    This abstraction may obscure finer-grained variations within each category.
    For instance, VQA covers diverse sub-skills such as counting, spatial reasoning, and commonsense QA, which may rely on richer and more distinct neuron-level mechanisms.
    Future work can adopt a finer task taxonomy to reveal sub-task-level sharing and conflict patterns.

    \item \textbf{Computational cost of causal analysis.}
    Our coarse-to-fine pipeline requires multiple forward passes, including targeted head ablations and verification through neuron-group and cross-task ablations.
    Scaling to more layers, more neurons, or larger datasets can therefore incur substantial computational overhead.
    Future work may explore more efficient candidate screening and approximate evaluation strategies to reduce inference-time cost.
\end{enumerate}

\bibliography{custom}

\appendix
\section{Benchmark Details}
\label{app:dataset-details}

\begin{table*}[t]
\centering
\renewcommand{\arraystretch}{1.15}
\resizebox{0.95\textwidth}{!}{
\begin{tabular}{@{}l c c p{0.58\textwidth}@{}}
\toprule
\textbf{Split} & \textbf{\#Images} & \textbf{\#Instances per Task} & \textbf{Primary Usage} \\
\midrule
Discovery & 7,000 & 7,000 (VQA/OCR/Caption/Retrieval) & Head localization and neuron identification. \\
Development & 2,000 & 2,000 (VQA/OCR/Caption/Retrieval) & Hyperparameter selection (e.g., scaling factors). \\
Test & 3,000 & 3,000 (VQA/OCR/Caption/Retrieval) & Final ablation verification and steering evaluation. \\
\midrule
Total & 12,000 & 12,000 (VQA/OCR/Caption/Retrieval) & -- \\
\bottomrule
\end{tabular}
}
\caption{Unified benchmark split statistics. Each image yields one instance for each task via prompt switching, hence equal per-task counts in every split.}
\label{tab:data_split}
\end{table*}

In this section, we provide comprehensive details regarding the construction and specific configurations of our unified multi-task benchmark. We first outline the dataset composition and our strict splitting strategy to prevent information leakage. Next, we elaborate on task-specific implementation details, including visual prompting for OCR and candidate mining for retrieval. Finally, we specify the evaluation metrics used to quantify model performance across all tasks.

\subsection{Benchmark Composition \& Splitting}
We base our experiments on the \texttt{train2014} split of \textbf{MS COCO}~\citep{lin2014coco}. We first collect candidate images by taking the image overlap among:
\begin{itemize}
    \setlength\itemsep{0em}
    \item \textbf{COCO-Text}~\citep{veit2016cocotext} (scene-text annotations on COCO images, providing text bounding boxes and transcriptions)
    \item \textbf{MSCOCO Captions}~\citep{chen2015cococaptions} (human-written image captions; we use the standard 5-reference setting)
    \item \textbf{VQAv2}~\citep{goyal2017vqav2} (a balanced VQA benchmark with open-ended questions and human answers)
\end{itemize}

We then filter the candidates to retain only images with complete annotations required by our four-task setup, resulting in exactly \textbf{12{,}000} images. Crucially, for each image, we keep the visual input fixed and derive four task inputs (VQA/OCR/Caption/Retrieval) by \emph{only} changing the task instruction prompts. All questions and annotations are directly derived from the official datasets; we do not synthesize new labels.

Following prior neuron-analysis practice \cite{huo-etal-2024-mmneuron} that separates the data used for neuron identification from held-out evaluation, we strictly partition the 12,000 images into three disjoint splits: \textbf{7,000} for \emph{Discovery} (head localization and neuron identification), \textbf{2,000} for \emph{Development} (steering hyperparameter selection), and \textbf{3,000} for \emph{Test} (final reporting, ablations, and steering evaluation). All splits are image-disjoint to prevent leakage. Table~\ref{tab:data_split} details the statistics and specific usage of each split.

\begin{table*}[t]
\centering

\begin{minipage}{0.97\textwidth} 
\centering
\small
\renewcommand{\arraystretch}{1.12}
\setlength{\tabcolsep}{4pt}

\captionsetup{
  width=\linewidth,
  justification=raggedright,
  singlelinecheck=false
}

\begin{adjustbox}{max width=\linewidth}
\begin{tabular}{|>{\centering\arraybackslash}m{1.8cm}
                |>{\centering\arraybackslash}m{4.4cm}
                |>{\raggedright\arraybackslash}m{7.3cm}
                |>{\centering\arraybackslash}m{2.4cm}|}
\hline
\textbf{Task} & \textbf{Image Example} & \textbf{Question / Instruction} & \textbf{GT / Target} \\
\hline
VQA &
\includegraphics[width=\linewidth,height=3.1cm,keepaspectratio]{Appendix/A-Figure1.jpg} &
What kind of animal is this? &
cat \\
\hline
OCR &
\includegraphics[width=\linewidth,height=3.1cm,keepaspectratio]{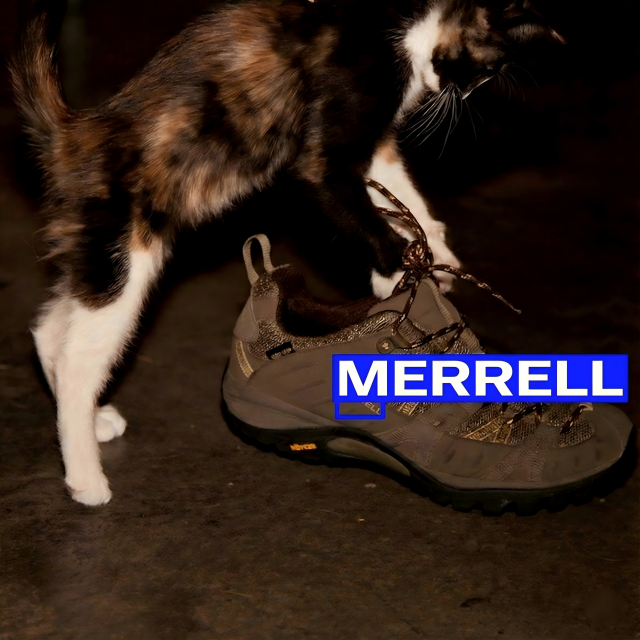} &
Transcribe the text inside the blue box. Output text only. &
MERRELL \\
\hline
Caption &
\includegraphics[width=\linewidth,height=3.1cm,keepaspectratio]{Appendix/A-Figure1.jpg} &
Describe the image in one sentence. &
A cat standing on top of a shoe on the floor. \\
\hline
Retrieval (I$\rightarrow$T) &
\includegraphics[width=\linewidth,height=3.1cm,keepaspectratio]{Appendix/A-Figure1.jpg} &
\makecell[l]{Does the candidate correctly describe the image? (Yes/No)\\
Text: [Candidate 1] A cat standing on top of a shoe.\\
Text: [Candidate 2] A dog running on the grass.\\
\textit{$\cdots$}\\
Text: [Candidate $N$] A group of people sitting in the room.} &
\makecell[c]{Yes\\No\\\textit{$\cdots$}\\No} \\

\hline
\end{tabular}
\end{adjustbox}

\caption{Task-specific examples in our unified benchmark. For OCR, the image is overlaid with a blue bounding box (from COCO-Text annotations) to indicate the target text region. Retrieval is evaluated as image-to-text (I$\rightarrow$T) pairwise verification and re-ranking.}
\label{tab:task_examples}
\end{minipage}
\end{table*}

\subsection{Task Implementation Details}

\paragraph{OCR (Visual Prompting).}
We leverage the scene-text bounding box annotations provided by \textbf{COCO-Text}. Using the annotated box coordinates, we overlay blue bounding boxes onto the original images to indicate the target text region, without cropping or changing image resolution. This visual prompt guides the model to the intended text instance for transcription in multi-text scenes.

\vspace{4pt}
\paragraph{Retrieval (Pairwise Re-ranking).}
We evaluate \emph{image-to-text} retrieval (I$\rightarrow$T): given an image query, the model selects the most relevant caption from a candidate pool. Since direct listwise ranking over a large candidate set is challenging for LVLMs, we reformulate I$\rightarrow$T retrieval as \emph{pairwise} image--text verification, where the model outputs ``Yes''/``No'' for each candidate text conditioned on the image, and we rank candidates by the probability of the token ``Yes''. To balance reliability and computational cost, we use a candidate pool size of 50 per query (1 positive + 49 negatives). For discovery-stage retrieval attribution, we construct candidates within the 7,000-image Discovery split as follows:
\begin{itemize}
    \setlength\itemsep{0em}
    \item \textbf{Query Set:} We use 2,000 images as queries. For each query image, we randomly sample \emph{one} caption from its five ground-truth captions as the positive candidate.
    \item \textbf{Hard Negative Pool:} We collect all captions from the remaining 5,000 images (5 captions per image; 25,000 captions in total) as a global negative pool, ensuring it is disjoint from the query set. Instead of randomly sampling negatives, we rank all pool captions by their \emph{semantic similarity} to the query (measured in a pretrained embedding space with cosine similarity), and select the top-49 most similar captions as negatives. This yields substantially harder and more realistic confounders for retrieval attribution.
\end{itemize}

\subsection{Evaluation Metrics}
We adopt standard metrics for each task:
\begin{itemize}
    \setlength\itemsep{0em}
    \item \textbf{VQA:} \textbf{VQAv2 Accuracy} computed with the official evaluation protocol (answer normalization and consensus-based soft scoring over 10 human answers).
    \item \textbf{OCR:} \textbf{ANLS}~\citep{Biten_2019_ICCV}, defined by normalized Levenshtein similarity with a threshold $\tau=0.5$.
    \item \textbf{Captioning:} \textbf{BLEU-4}~\citep{papineni2002bleu} using the standard COCO caption evaluation implementation with \emph{multiple references} (5 reference captions per image).
    \item \textbf{Retrieval:} \textbf{NDCG@5}~\citep{jarvelin2002cumulated} over the ranked captions for each image query:
    \[
    \begin{aligned}
    \mathrm{NDCG@5} &= \mathrm{DCG@5}/\mathrm{IDCG@5},\\
    \mathrm{DCG@5}  &= \sum_{k=1}^{5}\frac{2^{\mathrm{rel}_k}-1}{\log_2(k+1)}.
    \end{aligned}
    \]
    where $\mathrm{rel}_k$ denotes the relevance of the caption ranked at position $k$, and $\mathrm{IDCG@5}$ is the DCG@5 of the ideal (perfect) ranking.
\end{itemize}

\subsection{Task-Specific Examples}
Table~\ref{tab:task_examples} demonstrates representative inputs and targets for each task in our benchmark.

\section{Prompt Template}
\label{sec:prompt}

\subsection{Prompt Templates for Evaluation}
\label{app:prompt:evaluation}

To ensure consistent and interpretable input formatting across tasks and models, we design task-specific prompt templates for both \textbf{LLaVA-1.5-7B} and \textbf{Qwen2.5-VL-7B}. All templates follow the same principle: we keep the visual input fixed and \emph{only} vary the task instruction prompt to induce task-specific readouts. Below, \texttt{<Image>} denotes the image input, and \{\texttt{Question}\}/\{\texttt{CandidateText}\} are placeholders filled from the benchmark.

\begin{figure*}[t]
    \centering
    \includegraphics[width=\textwidth]{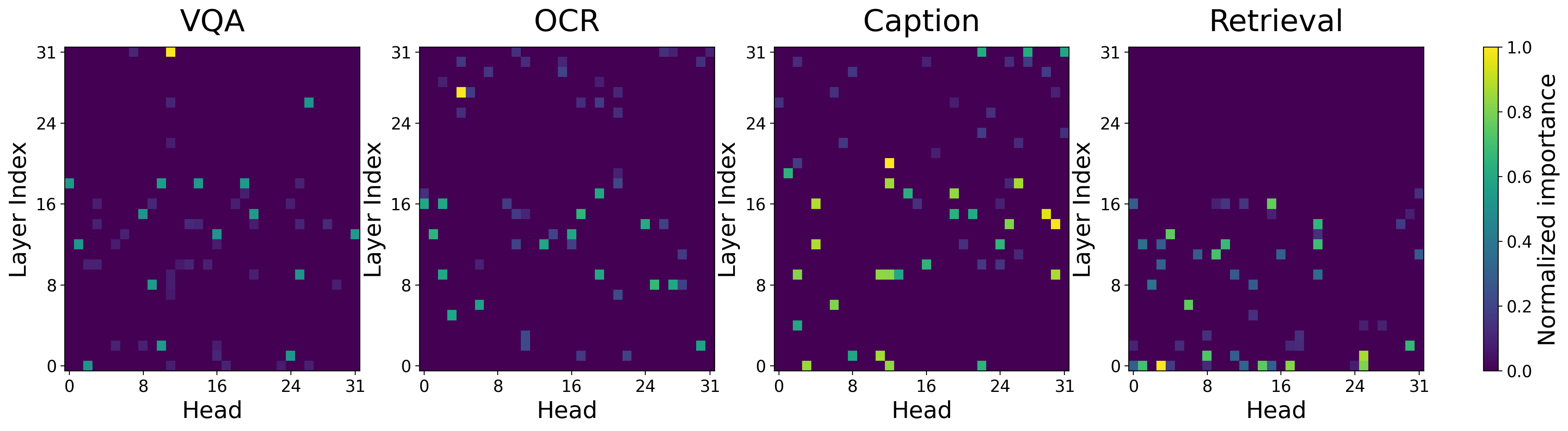}
    \caption{Attention head importance heatmaps for LLaVA-1.5-7B.
    Rows denote layer indices and columns denote head indices; colors indicate normalized head importance, with brighter colors meaning higher importance.}
    \label{fig:head_heatmaps_llava}
\end{figure*}

\begin{figure*}[t]
    \centering
    \includegraphics[width=\textwidth]{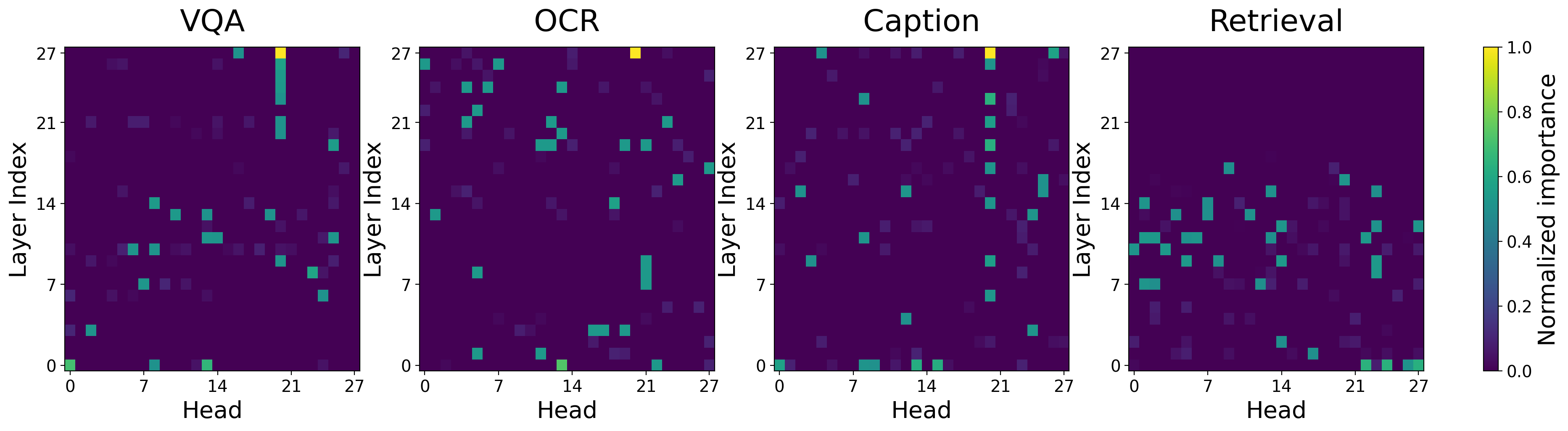}
    \caption{Attention head importance heatmaps for Qwen2.5-VL-7B.
    Rows denote layer indices and columns denote head indices; colors indicate normalized head importance, with brighter colors meaning higher importance.}
    \label{fig:head_heatmaps_qwen}
\end{figure*}

\section{Task-Critical Attention Heads}
\label{sec:finding-heads}

In this section, we visualize the spatial distribution of task-critical attention heads to elucidate the routing mechanisms within VLMs.
Using the head-level causal localization procedure detailed in Sec.~\ref{sec:head-local} (referencing V-SEAM \citep{wang-etal-2025-v}), we compute the normalized importance of each head across four tasks (VQA, OCR, Captioning, Retrieval) for both LLaVA-1.5-7B and Qwen2.5-VL-7B. The resulting heatmaps are presented in Fig.~\ref{fig:head_heatmaps_llava} and Fig.~\ref{fig:head_heatmaps_qwen}.

Across models, we observe three consistent patterns.
(i) \textbf{Sparsity:} Important heads are highly sparse, indicating that task-relevant information flow is dominated by a small set of routing hubs.
(ii) \textbf{A shared mid-layer backbone:} Both models exhibit a higher density of critical heads in mid layers, suggesting a shared backbone where cross-modal interactions are most concentrated.
(iii) \textbf{Task-dependent depth profiles:} Beyond the shared backbone, different tasks exhibit distinct depth patterns: Retrieval tends to emphasize early-to-mid layers, while OCR and Captioning assign more weight to mid-to-late layers; VQA shows a more distributed profile across depth.

These observations motivate the \emph{coarse-to-fine} design of \textbf{HONES}: since computation is routed through sparse, task-specific pathways, we condition neuron attribution on the localized routing heads (see \S\ref{sec:neuron-attrib}).

\begin{figure*}[t]
    \centering
    \includegraphics[width=0.49\textwidth]{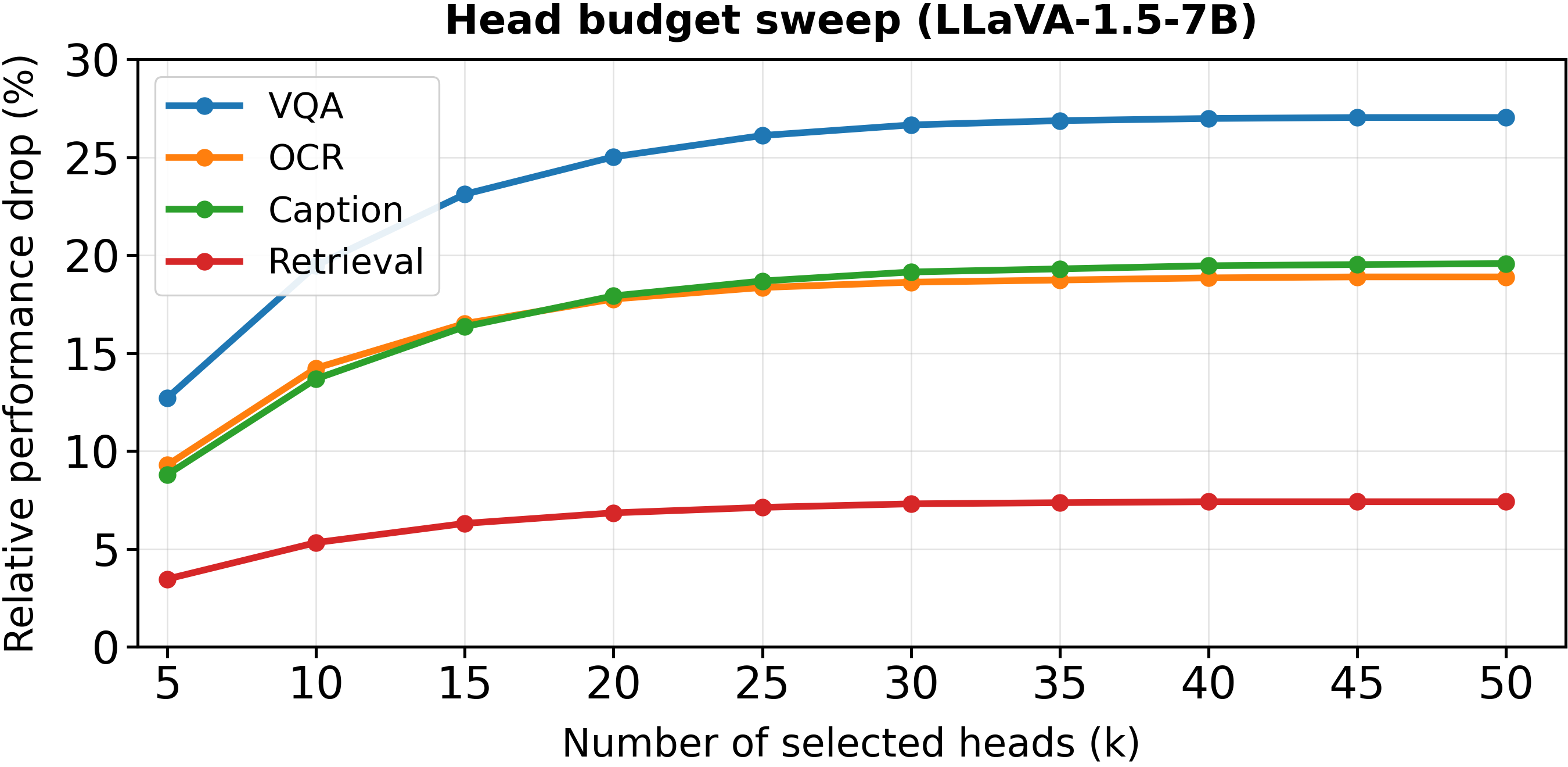}
    \hfill
    \includegraphics[width=0.49\textwidth]{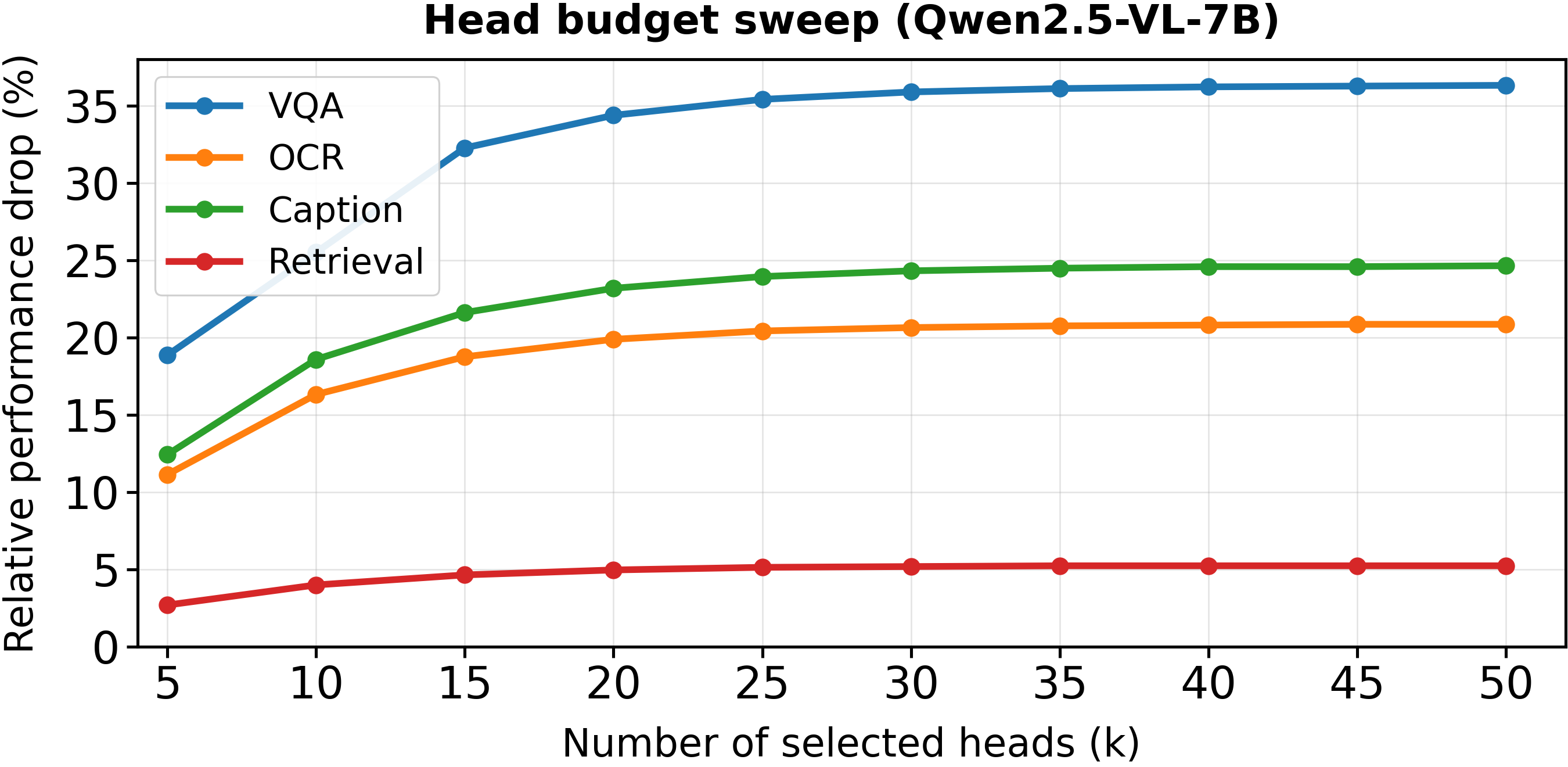}
    \caption{Attention head budget sweep. Relative performance drop (\%) after masking the top-$K$ localized heads across four tasks.
    The curves exhibit clear saturation points, motivating $K{=}30$ for LLaVA-1.5-7B and $K{=}25$ for Qwen2.5-VL-7B to capture the sparse routing backbone with diminishing returns beyond the elbow.}
    \label{fig:head_budget_ablation}
    \vspace{-1mm}
\end{figure*}

\section{Additional Implementation Details and Results for Neuron Localization}
\label{sec:appendix-neuron-details}

In this section, we provide supplementary details regarding the neuron localization experiments presented in the main text (\S\ref{sec:key-findings}), including the hyperparameter selection for attention heads, formal definitions of baseline strategies, and the absolute performance metrics corresponding to the relative drops reported in Table~\ref{tab:neuron_main_two_models}, as well as additional larger-backbone validation and efficiency analysis.

\subsection{IDF Weight.}
\label{sec:IDF}
We compute $\alpha(\tau)$ on the discovery split.
Let $N$ be the number of instances in $\mathcal{D}_t^{\mathrm{disc}}$ and
$\mathrm{df}(\tau)$ be the number of instances whose references contain token $\tau$.
We use the smoothed IDF:
\begin{equation}
\alpha(\tau) = \log\frac{N+1}{\mathrm{df}(\tau)+1} + 1.
\end{equation}

Given $\mathbf{u}_{\mathrm{tgt}}(x,y)$, we define the \emph{sample-level} write-in score for run $\rho$ as
\begin{equation}
\label{eq:writein_score}
c^{\rho}_{\ell,i}(x,y)
=
\left\langle \Delta \mathbf{r}^{(\ell),\rho}_i(x),\, \mathbf{u}_{\mathrm{tgt}}(x,y)\right\rangle.
\end{equation}

\subsection{Ablation Study on Attention Head Budget}
\label{sec:head-budget-ablation}

To determine the optimal head budget $K_h$ for constraining the neuron search space in \textsc{HONES}, 
we conduct a sensitivity sweep over $K_h \in \{5, 10, \dots, 50\}$ by measuring the relative performance drop
induced by masking the top-$K_h$ localized heads for each task.
Our selection criterion is based on identifying the saturation point (or ``elbow'') of the performance curves:
a steep initial rise indicates that the top-ranked heads capture sparse routing hubs effectively, 
whereas a subsequent plateau suggests diminishing returns---including additional heads beyond this point
contributes only marginally to task routing and may introduce noise into the downstream neuron attribution.

Fig.~\ref{fig:head_budget_ablation} summarizes the trends for both models.
For \textbf{LLaVA-1.5-7B}, the curves increase rapidly and then saturate at task-dependent elbows:
VQA saturates around $K_h{=}28$, OCR around $K_h{=}26$, Caption around $K_h{=}30$, and Retrieval around $K_h{=}28$.
To ensure consistent coverage across tasks while staying near the saturation regime, we adopt a unified budget of $\mathbf{K_h{=}30}$ for LLaVA-1.5-7B.
For \textbf{Qwen2.5-VL-7B}, saturation happens earlier:
VQA around $K_h{=}24$, OCR around $K_h{=}23$, Caption around $K_h{=}25$, and Retrieval around $K_h{=}24$,
thus we set $\mathbf{K_h{=}25}$ for Qwen.

Finally, note that masking far more heads (e.g., $K_h \gg 50$) would likely continue to degrade performance and may eventually cause collapse; however, our goal here is \emph{not} to maximize degradation, but to identify a \emph{sparse} and \emph{causally-relevant} routing backbone that stabilizes the neuron search space.
Selecting $K_h$ at the plateau balances two factors: (i) capturing the dominant information flow and (ii) avoiding the inclusion of weakly-related heads that could dilute head-constrained neuron attribution.

\begin{table*}[t]
\centering
\scriptsize
\renewcommand{\arraystretch}{1.12}
\setlength{\tabcolsep}{2.8pt}
\resizebox{\textwidth}{!}{
\begin{tabular}{@{}lccccc@{\hspace{0.7em}}ccccc@{}}
\toprule
\multirow{2}{*}{\textbf{Strategy}} &
\multicolumn{5}{c}{\textbf{LLaVA-1.5-7B}} &
\multicolumn{5}{c}{\textbf{Qwen2.5-VL-7B}} \\
\cmidrule(lr){2-6}\cmidrule(lr){7-11}
& \textbf{VQA} & \textbf{OCR} & \textbf{Caption} & \textbf{Retrieval} & \textbf{Avg}
& \textbf{VQA} & \textbf{OCR} & \textbf{Caption} & \textbf{Retrieval} & \textbf{Avg} \\
\midrule
Base VLM (unmasked)              & 0.6520 & 0.5760 & 0.1285 & 0.9328 & 0.5723 & 0.6750 & 0.6580 & 0.2240 & 0.9464 & 0.6259 \\
\midrule
AP \cite{huang2024miner}          & 0.5780 & 0.5161 & 0.1174 & 0.9281 & 0.5349 & 0.6410 & 0.5896 & 0.1915 & 0.9446 & 0.5917 \\
MA \cite{xu2025deciphering}             & 0.6073 & 0.4867 & 0.1132 & 0.9202 & 0.5319 & 0.5260 & 0.5560 & 0.2019 & 0.9402 & 0.5561 \\
APE \cite{huo-etal-2024-mmneuron}    & 0.6310 & 0.5868 & 0.1128 & 0.9244 & 0.5638 & 0.6907 & 0.6703 & 0.2101 & 0.9456 & 0.6292 \\
\midrule
HONES-RandHead          & 0.5887 & 0.5397 & 0.1157 & 0.9188 & 0.5407 & 0.5873 & 0.5823 & 0.1837 & 0.9109 & 0.5660 \\
HONES-Gaussian        & 0.6227 & 0.5553 & 0.1164 & 0.9239 & 0.5546 & 0.6377 & 0.5988 & 0.1960 & 0.9388 & 0.5929 \\
RandNeuron             & 0.6463 & 0.5667 & 0.1255 & 0.9322 & 0.5677 & 0.6587 & 0.6450 & 0.2166 & 0.9450 & 0.6163 \\
\midrule
\textbf{HONES (Ours)}              & 0.4740 & 0.4666 & 0.1031 & 0.8635 & 0.4768 & 0.4287 & 0.5198 & 0.1684 & 0.8958 & 0.5032 \\
\bottomrule
\end{tabular}
}
\caption{Absolute performance after masking top-1\% neurons. 
Avg is the arithmetic mean of the four task scores.}
\label{tab:neuron_main_two_models_abs}
\vspace{-2mm}
\end{table*}

\begin{table}[t]
\centering
\scriptsize
\renewcommand{\arraystretch}{1.1}
\setlength{\tabcolsep}{4pt}
\resizebox{\columnwidth}{!}{%
\begin{tabular}{lccccc}
\toprule
\textbf{Method} & \textbf{VQA} & \textbf{OCR} & \textbf{Caption} & \textbf{Retrieval} & \textbf{Avg} \\
\midrule
AP~\mbox{\citep{huang2024miner}}            & 12.45 &  9.30 &  9.50 & 0.70 &  7.99 \\
MA~\mbox{\citep{xu2025deciphering}}         &  5.35 & 16.25 & 13.40 & 1.65 &  9.16 \\
APE~\mbox{\citep{huo-etal-2024-mmneuron}}   &  3.90 &  2.80 & 12.90 & 1.40 &  5.25 \\
\midrule
HONES-RandHead   & 13.25 &  7.60 & 11.40 & 1.60 &  8.46 \\
HONES-Gaussian   &  7.10 &  4.50 &  9.90 & 1.10 &  5.65 \\
RandNeuron       &  0.95 &  1.80 &  2.05 & 0.15 &  1.24 \\
\midrule
\textbf{HONES (Ours)} & \textbf{24.15} & \textbf{21.30} & \textbf{26.00} & \textbf{5.80} & \textbf{19.31} \\
\bottomrule
\end{tabular}%
}
\caption{Additional scale validation on \textbf{LLaVA-1.5-13B}. Relative performance drop (\%) after masking the top-1\% neurons selected by each ranking method. Higher values indicate greater causal importance (negative values indicate improvement). Avg denotes the arithmetic mean across the four tasks.}
\label{tab:llava13b_scale}
\vspace{-4mm}
\end{table}

\subsection{Definitions of Baseline Strategies}
\label{sec:baseline-def}
To rigorously evaluate the efficacy of \textsc{HONES}, we compare it against three widely used activation-based heuristics and a random baseline.
Let $a^{l}_{ij}(x)$ denote the post-activation output (e.g., after GeLU) of neuron $j$ at layer $l$ on token position $i$ for input $x$.
Since VLM inputs contain multiple tokens (visual and textual), we first aggregate activations over positions to obtain a single scalar per neuron:
\[
\tilde{a}^{l}_{j}(x) \;=\; \max_{i \in \mathcal{I}(x)} a^{l}_{ij}(x),
\]
where $\mathcal{I}(x)$ denotes the set of token positions for $x$.
All statistics below are computed on the same discovery split used for neuron selection, and we always select the top-$B$ neurons under each strategy with an identical budget $B$ (Top-1\% of all MLP neurons).

\paragraph{Activation Probability (AP; ``activation frequency'').}
This heuristic ranks neurons by how often they are activated across the dataset, a common proxy used in modality-/domain-sensitive neuron discovery.
Formally, given a task-specific dataset $\mathcal{D}$ and an activation threshold $\tau$ (we set $\tau = 0$), we define
\[
p^{l}_{j} \;=\; \frac{1}{|\mathcal{D}|}\sum_{x \in \mathcal{D}} \mathbb{I}\!\left[\tilde{a}^{l}_{j}(x) > \tau\right].
\]
Neurons are then ranked by $p^{l}_{j}$ in descending order, and the top-$B$ are selected.
This baseline is aligned with activation-probability-based neuron mining in multimodal settings (e.g., \citealp{huang2024miner}).

\paragraph{Mean Activation Magnitude (MA; ``mean activation rank'').}
Instead of frequency, this baseline ranks neurons by their average activation strength, motivated by the intuition that consistently large activations may indicate frequent usage.
We compute
\[
\mu^{l}_{j} \;=\; \frac{1}{|\mathcal{D}|}\sum_{x \in \mathcal{D}} \left|\tilde{a}^{l}_{j}(x)\right|,
\]
and rank neurons by $\mu^{l}_{j}$ in descending order.
This style of magnitude-based scoring is commonly adopted when probing functional neurons in VLMs (e.g., \citealp{xu2025deciphering}).

\paragraph{Activation Probability Entropy (APE; ``activation probability entropy'').}
Entropy-based selection aims to capture neurons with uncertain on/off patterns across samples, which has been used as a heuristic signal for identifying neurons that respond to diverse inputs.
We compute the Bernoulli entropy of $p^{l}_{j}$:
\[
H^{l}_{j} \;=\; -p^{l}_{j}\log(p^{l}_{j}) \;-\; (1-p^{l}_{j})\log(1-p^{l}_{j}),
\]
and rank neurons by $H^{l}_{j}$ in descending order.
This baseline follows probability-entropy-style scoring used in multimodal neuron analysis (e.g., \citealp{huo-etal-2024-mmneuron}).

\paragraph{Random Selection.}
As a lower-bound baseline, we uniformly sample $B$ neurons from the full set of MLP neurons (across all layers), repeat this procedure for multiple seeds, and report the mean performance drop.
This controls for the neuron budget and quantifies the expected effect of masking arbitrary neurons.

\subsection{Absolute Performance Metrics}
\label{sec:absolute-metrics}
Table~\ref{tab:neuron_main_two_models_abs} reports the \emph{absolute} performance scores (Accuracy for VQA, ANLS for OCR, BLEU-4 for Captioning, and NDCG@5 for Retrieval) after masking the top-1\% neurons identified by different strategies.
These raw metrics correspond to the relative drops reported in the main text.
Consistent with the relative analysis, \textbf{HONES} yields the lowest absolute scores (i.e., the largest damage) across tasks, confirming its effectiveness in localizing causally critical neurons.

\begin{table}[t]
\centering
\footnotesize
\renewcommand{\arraystretch}{1.10}
\setlength{\tabcolsep}{3pt}
\resizebox{\columnwidth}{!}{%
\begin{tabular}{llc}
\toprule
\textbf{Model} & \textbf{Method} & \textbf{Time (s/inst.)}$\downarrow$ \\
\midrule
LLaVA-1.5-7B & Group Patching~\citep{zhang2023towards} & 82.40 $\pm$ 5.60 \\
LLaVA-1.5-7B & QRNCA~\citep{chen2025identifying} & 46.50 $\pm$ 3.45 \\
LLaVA-1.5-7B & \textbf{HONES} & \textbf{8.30 $\pm$ 0.50} \\
\midrule
Qwen2.5-VL-7B & Group Patching~\citep{zhang2023towards} & 106.20 $\pm$ 7.90 \\
Qwen2.5-VL-7B & QRNCA~\citep{chen2025identifying} & 53.80 $\pm$ 3.90 \\
Qwen2.5-VL-7B & \textbf{HONES} & \textbf{10.15 $\pm$ 0.65} \\
\bottomrule
\end{tabular}%
}
\caption{Efficiency assessment of neuron localization on VQA. \textbf{Time} reports the average localization runtime per instance (s/inst.) under the same hardware environment. All numbers are reported as mean $\pm$ standard deviation over 10 random trials.}
\label{tab:efficiency_compare}
\vspace{-6mm}
\end{table}

\begin{table*}[t]
\centering
\small
\setlength{\tabcolsep}{6pt}
\renewcommand{\arraystretch}{1.15}

\begin{tabular*}{\textwidth}{@{\extracolsep{\fill}} l
S[table-format=-2.2] S[table-format=-2.2] S[table-format=-2.2] S[table-format=-2.2]}
\toprule
\textbf{Masked neuron group} &
{\textbf{VQA}} & {\textbf{OCR}} & {\textbf{Caption}} & {\textbf{Retrieval}} \\
\midrule

\multicolumn{5}{@{}l}{\textit{Specific (only one task)}} \\
VQA-specific        &  5.90 &  1.50 &  1.50 &  1.30 \\
OCR-specific        & -0.50 &  6.50 & -0.40 & -0.60 \\
Caption-specific    &  0.50 & -2.00 &  6.20 & -0.20 \\
Retrieval-specific  &  0.30 & -2.30 & -2.70 &  2.72 \\
\midrule

\multicolumn{5}{@{}l}{\textit{Pair (two-task shared)}} \\
VQA+OCR           &  1.30 & \best{9.80} & \multicolumn{1}{c}{--} & \multicolumn{1}{c}{--} \\
VQA+Caption       &  0.50 & \multicolumn{1}{c}{--} &  5.00 & \multicolumn{1}{c}{--} \\
VQA+Retrieval     & \best{10.70} & \multicolumn{1}{c}{--} & \multicolumn{1}{c}{--} & \best{6.30} \\
OCR+Caption       & \multicolumn{1}{c}{--} & -1.20 &  0.90 & \multicolumn{1}{c}{--} \\
OCR+Retrieval     & \multicolumn{1}{c}{--} &  2.00 & \multicolumn{1}{c}{--} &  0.40 \\
Caption+Retrieval & \multicolumn{1}{c}{--} & \multicolumn{1}{c}{--} & -2.80 & -0.50 \\
\midrule

\multicolumn{5}{@{}l}{\textit{Triple (three-task shared)}} \\
VQA+OCR+Caption       &  1.40 &  3.10 & \best{14.30} & \multicolumn{1}{c}{--} \\
VQA+OCR+Retrieval     &  1.00 &  4.50 & \multicolumn{1}{c}{--} &  0.60 \\
VQA+Caption+Retrieval &  0.40 & \multicolumn{1}{c}{--} &  0.90 & -1.00 \\
OCR+Caption+Retrieval & \multicolumn{1}{c}{--} & -1.20 & -0.60 &  0.20 \\
\midrule

\multicolumn{5}{@{}l}{\textit{General (four-task shared)}} \\
GENERAL & 1.40 & 0.20 & 3.90 & 1.20 \\
\midrule

\multicolumn{5}{@{}l}{\makecell[l]{\textit{Random control (matched to the most critical shared group;}\\
\textit{mean of 10 trials)}}} \\
Random (same size) & 0.20 & 0.80 & -1.10 & 0.10 \\
\bottomrule
\end{tabular*}

\caption{Cross-task neuron ablation results on LLaVA-1.5.
Values report relative performance change (\%); positive indicates performance drop, while negative indicates performance gain.
For each target task, we bold the single most damaging \emph{shared-group} ablation (Pair/Triple/General), excluding task-specific (\emph{Only}) and random controls.
Random controls are computed by masking the same number of neurons as the most critical shared group for each target task (mean over 10 trials).}
\label{tab:llava_cross_task_ablation_full}
\end{table*}

\begin{table*}[t]
\centering
\small
\setlength{\tabcolsep}{6pt}
\renewcommand{\arraystretch}{1.15}

\begin{tabular*}{\textwidth}{@{\extracolsep{\fill}} l
S[table-format=-2.2] S[table-format=-2.2] S[table-format=-2.2] S[table-format=-2.2]}
\toprule
\textbf{Masked neuron group} &
{\textbf{VQA}} & {\textbf{OCR}} & {\textbf{Caption}} & {\textbf{Retrieval}} \\
\midrule

\multicolumn{5}{@{}l}{\textit{Specific (only one task)}} \\
VQA-specific        & 14.80 &  1.90 &  2.10 &  0.60 \\
OCR-specific        & -3.40 &  9.20 & -0.90 & -0.25 \\
Caption-specific    &  2.35 & -3.80 &  7.35 &  0.10 \\
Retrieval-specific  &  1.95 & -2.70 & -2.90 &  1.80 \\
\midrule

\multicolumn{5}{@{}l}{\textit{Pair (two-task shared)}} \\
VQA+OCR           &  3.90 & \best{12.50} & \multicolumn{1}{c}{--} & \multicolumn{1}{c}{--} \\
VQA+Caption       &  2.10 & \multicolumn{1}{c}{--} &  6.10 & \multicolumn{1}{c}{--} \\
VQA+Retrieval     & \best{26.01} & \multicolumn{1}{c}{--} & \multicolumn{1}{c}{--} & \best{3.41} \\
OCR+Caption       & \multicolumn{1}{c}{--} & -2.30 & -1.30 & \multicolumn{1}{c}{--} \\
OCR+Retrieval     & \multicolumn{1}{c}{--} &  2.70 & \multicolumn{1}{c}{--} &  0.75 \\
Caption+Retrieval & \multicolumn{1}{c}{--} & \multicolumn{1}{c}{--} &  1.60 & -0.20 \\
\midrule

\multicolumn{5}{@{}l}{\textit{Triple (three-task shared)}} \\
VQA+OCR+Caption       &  4.80 &  3.80 & \best{16.12} & \multicolumn{1}{c}{--} \\
VQA+OCR+Retrieval     &  5.20 &  5.80 & \multicolumn{1}{c}{--} &  0.90 \\
VQA+Caption+Retrieval &  3.40 & \multicolumn{1}{c}{--} &  3.00 &  0.35 \\
OCR+Caption+Retrieval & \multicolumn{1}{c}{--} & -2.60 & -2.20 &  0.15 \\
\midrule

\multicolumn{5}{@{}l}{\textit{General (four-task shared)}} \\
GENERAL & 4.10 & 3.90 & 5.25 & 0.95 \\
\midrule

\multicolumn{5}{@{}l}{\makecell[l]{\textit{Random control (matched to the most critical shared group;}\\
\textit{mean of 10 trials)}}} \\
Random (same size) & 1.25 & 1.70 & 2.25 & 0.38 \\
\bottomrule
\end{tabular*}

\caption{Cross-task neuron ablation results on Qwen2.5-VL.
Values report relative performance change (\%); positive indicates performance drop, while negative indicates performance gain.
For each target task, we bold the single most damaging \emph{shared-group} ablation (Pair/Triple/General), excluding task-specific (\emph{Only}) and random controls.
Random controls are computed by masking the same number of neurons as the most critical shared group for each target task (mean over 10 trials).}
\label{tab:qwen_cross_task_ablation_s}
\end{table*}

\begin{figure*}[!t]
    \centering

    \begin{subfigure}[t]{0.96\textwidth}
        \centering
        \includegraphics[width=\textwidth]{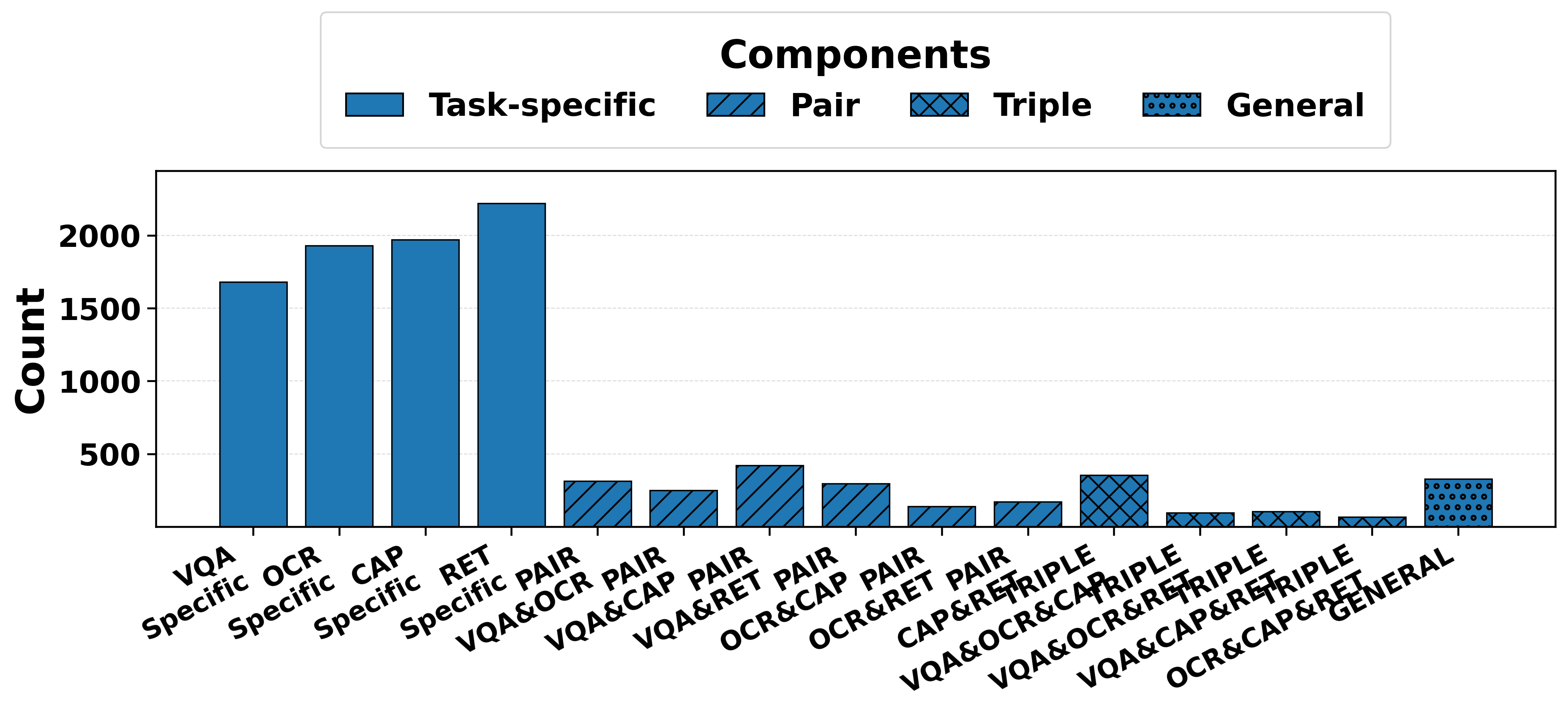}
        \caption{Neuron overlap composition across tasks (\textsc{LLaVA}-1.5).}
        \label{fig:appendix_neuron_overlap_counts_llava}
    \end{subfigure}

    \vspace{2mm}

    \begin{subfigure}[t]{0.96\textwidth}
        \centering
        \includegraphics[width=\textwidth]{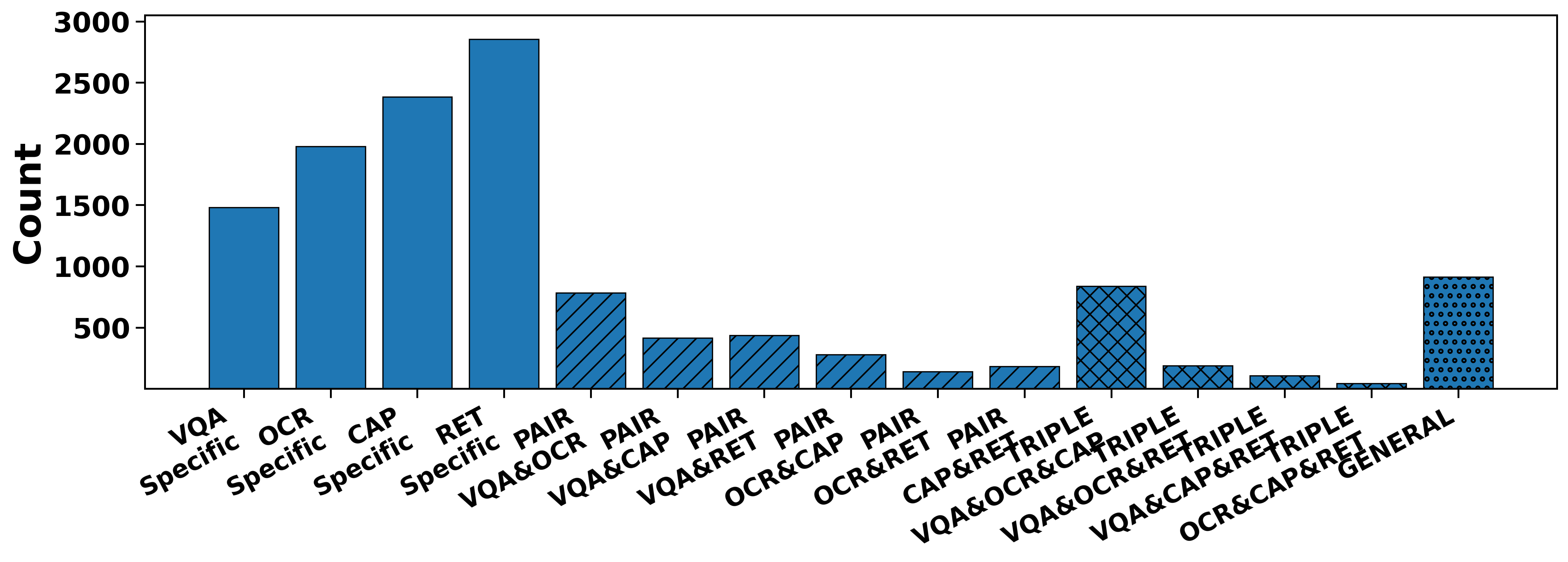}
        \caption{Neuron overlap composition across tasks (\textsc{Qwen2.5-VL}).}
        \label{fig:appendix_neuron_overlap_counts_qwen}
    \end{subfigure}

    \caption{Neuron overlap composition across tasks. Counts of task-critical neurons are partitioned into 15 mutually exclusive overlap groups induced by four tasks (VQA/OCR/Caption/Retrieval), including task-specific, pair, triple, and general components.}
    \label{fig:appendix_neuron_overlap_counts}
\end{figure*}

\subsection{Additional Scale Validation on a Larger Backbone}
\label{app:llava13b}

To assess whether our neuron localization results remain stable beyond 7B dense backbones, we further evaluate HONES on a larger model, \textbf{LLaVA-1.5-13B}, under the same four-task setting as in the main paper. The results remain consistent with those observed in the 7B models, with HONES continuing to be the strongest localization method across all four tasks. Table~\ref{tab:llava13b_scale} shows the results.

\subsection{Efficiency Assessment on VQA}
\label{sec:appendix_efficiency}

To further quantify the computational cost of neuron localization, we compare \textbf{HONES} with two representative baselines under a unified VQA evaluation protocol: the gradient-based attribution method \textbf{QRNCA}~\citep{chen2025identifying} and the explicit intervention-based method \textbf{Group Patching}~\citep{zhang2023towards}. We choose these two methods because they represent the two types of comparison targets most relevant to HONES: one is gradient-based methods, which are used to examine the efficiency advantage of HONES as a gradient-free method; the other is causal intervention-based localization methods, which are used to compare the efficiency of HONES against strong intervention-style attribution methods. Meanwhile, in the preceding comparison of localization effectiveness, these two types of methods are also among the most competitive non-HONES baselines. Specifically, we conduct 10 independent random trials on the VQA analysis split, where each trial samples 200 examples for neuron ranking and localization. For fairness, all methods use the same candidate space (all FFN neurons), identical data preprocessing, and the same masking budget (top-1\% neurons) throughout the comparison.

Under the same hardware environment (Tesla V100 GPU and Intel Xeon Gold 6230R CPU), we report the average localization runtime per instance for each method. As shown in Table~\ref{tab:efficiency_compare}, on LLaVA-1.5-7B and Qwen2.5-VL-7B, the per-instance localization time of HONES is consistently much lower than that of the two baselines, reaching 8.30$\pm$0.50s and 10.15$\pm$0.65s, respectively; correspondingly, QRNCA reaches 46.50$\pm$3.45s and 53.80$\pm$3.90s, and Group Patching reaches 82.40$\pm$5.60s and 106.20$\pm$7.90s. These results indicate that, under the same evaluation protocol, HONES achieves substantially higher neuron-localization efficiency than the compared baselines.

\section{Supplementary Cross-Task Neuron Ablation Results}
\label{app:supp_cross_task_neuron_ablation}

\subsection{Additional information for Cross-Task Neuron Analysis}
\label{sec:appendix_finding3}

\paragraph{Full cross-task ablations and group definitions.}
We provide the complete cross-task neuron ablation tables for LLaVA-1.5 and Qwen2.5-VL in
Table~\ref{tab:llava_cross_task_ablation_full} and Table~\ref{tab:qwen_cross_task_ablation_s}.
All neuron groups are constructed from the \textbf{top-1\%} task-critical neuron sets $\mathcal{N}_t$ identified by \textbf{HONES} (ranked by \emph{Causal Write-in Effect}).
Concretely, let $\mathcal{N}_t$ denote the top-1\% task-critical neuron set for task $t\in\{\text{VQA},\text{OCR},\text{Cap},\text{Ret}\}$.
We partition neurons into four categories by set union/intersection:
(1) \emph{Task-specific}: neurons belonging to exactly one task set (e.g., $\mathcal{N}_{\text{VQA}}\setminus(\mathcal{N}_{\text{OCR}}\cup\mathcal{N}_{\text{Cap}}\cup\mathcal{N}_{\text{Ret}})$);
(2) \emph{Pair}: neurons shared by \emph{exactly} two tasks (e.g., $(\mathcal{N}_{\text{VQA}}\cap\mathcal{N}_{\text{Ret}})\setminus \mathcal{N}_{\text{OCR}}\setminus \mathcal{N}_{\text{Cap}}$);
(3) \emph{Triple}: neurons shared by \emph{exactly} three tasks;
(4) \emph{General}: neurons shared by all four tasks ($\mathcal{N}_{\text{VQA}}\cap\mathcal{N}_{\text{OCR}}\cap\mathcal{N}_{\text{Cap}}\cap\mathcal{N}_{\text{Ret}}$).
For each target task, we highlight the single most damaging \emph{shared-group} ablation among Pair/Triple/\emph{General} to define the \emph{dominant salient neuron group} (as summarized in Fig.~\ref{fig:cross_task_ablation_bars}).

\paragraph{Neuron group composition.}
As shown in Fig.~\ref{fig:appendix_neuron_overlap_counts_llava} and Fig.~\ref{fig:appendix_neuron_overlap_counts_qwen}, task-critical neurons are first identified \emph{separately} for each of the four tasks by selecting the top-1\% neurons.
We then jointly consider the four task-wise top-1\% sets and assign each neuron to an overlap group based on its cross-task membership pattern, yielding 15 mutually exclusive groups (4 task-specific, 6 pair, 4 triple, and 1 general).
Across both backbones, the distribution is highly skewed: task-specific neurons account for the majority of critical units, while higher-order overlaps (pair/triple/general) form a much smaller fraction.
Moreover, shared neurons are not uniformly spread across combinations; instead, they concentrate on a few pairs/triples that involve VQA, consistent with sparse and structured sharing rather than a dense universal subnetwork.

\paragraph{Additional evidence for VQA-centric selective sharing.}
Tables~\ref{tab:llava_cross_task_ablation_full} and \ref{tab:qwen_cross_task_ablation_s} further support the main-text conclusion that cross-task reuse is \emph{sparse and structured}, rather than uniformly distributed.
First, across both backbones, the dominant salient neuron group for every target task always overlaps with VQA (e.g., \emph{VQA+Ret} for Retrieval and VQA, \emph{VQA+OCR} for OCR, and \emph{VQA+OCR+Cap} for Caption), consistent with a VQA-centered hub.
Second, the four-task \emph{General} group is \emph{not} the most damaging shared set, indicating that multi-task transfer is dominated by \emph{combination-specific} bridges rather than a single universal subnetwork.
Third, several non-VQA shared groups yield marginal or even negative drops, suggesting mixed-use computation and potential cross-task interference; this evidence supports our head-constrained, readout-aligned neuron discovery in \textbf{HONES}, which narrows the search to task-relevant routing pathways and thus better isolates verifiable causal units for reliable intervention.

\paragraph{Ruling out group-size artifacts via matched random ablation.}
Tables~\ref{tab:llava_cross_task_ablation_full} and \ref{tab:qwen_cross_task_ablation_s} also report matched random ablation:
for each target task, we mask a random neuron set with the \emph{same size} as the most critical shared group (mean over 10 trials).
The resulting drops are consistently much smaller than those caused by the dominant salient neuron group, ruling out a trivial ``mask-more-drop-more'' explanation and supporting a structured causal dependency.
This further implies that causally important neurons are sparsely and non-uniformly distributed, with their density varying across depth and tasks, rather than being evenly spread across neurons, suggesting that a small subset of units carries a disproportionate share of task-critical computation in multi-task VLMs.

\subsection{Exploring Text-Cue Dependence in Captioning via Visual-Text Masking}
\label{sec:appendix_caption_text_control}

\paragraph{Motivation and setup.}
In the main experiments, Captioning can be influenced by explicit visual-text cues (e.g., scene text present in images).
To test whether the \emph{Triple} bridge (\emph{VQA+OCR+Cap}) is driven by such cues, we perform a text-factor control study:
we \emph{mask out visual text regions} during inference for the Caption task, while keeping the rest of the pipeline unchanged.
Concretely, we use the COCO-Text annotations (character/word bounding boxes) to overwrite text pixels with a neutral patch (e.g., mean-color) on the input image before feeding it to the model.

\paragraph{Key observation: the shared bridge contracts from Triple to Pair.}
As shown in Table~\ref{tab:caption_text_control}, after removing explicit text cues, the dominant salient neuron group for Caption consistently shifts from \emph{VQA+OCR+Cap} to \emph{VQA+Cap} on both backbones,
indicating that the previously observed Triple bridge is largely triggered by OCR-relevant computation induced by scene text.
This provides direct causal support for the interpretation in the main text: Captioning couples \emph{text understanding} (OCR) and \emph{text generation} only when explicit text cues are present; otherwise it behaves as a more standard vision-to-text generation task that primarily shares with VQA-style semantic understanding.

\begin{table}[t]
\centering
\small
\setlength{\tabcolsep}{6pt}
\renewcommand{\arraystretch}{1.15}
\begin{tabular}{l|l|c}
\toprule
\textbf{Setting} & \textbf{Masked bridge} & \textbf{Caption drop (\%)} \\
\midrule
\multicolumn{3}{l}{\textit{LLaVA-1.5-7B}} \\
Original     & VQA+OCR+Cap & 14.30 \\
Original     & VQA+Cap     & 5.00  \\
Text-masked  & VQA+OCR+Cap & 4.37  \\
Text-masked  & VQA+Cap     & 11.00 \\
\midrule
\multicolumn{3}{l}{\textit{Qwen2.5-VL-7B}} \\
Original     & VQA+OCR+Cap & 16.12 \\
Original     & VQA+Cap     & 6.10  \\
Text-masked  & VQA+OCR+Cap & 7.68  \\
Text-masked  & VQA+Cap     & 13.25 \\
\bottomrule
\end{tabular}
\caption{Caption text-factor control.
We compare two candidate shared bridges (Triple: VQA+OCR+Cap; Pair: VQA+Cap) under the original setting and the text-masked setting.
After masking explicit visual text cues, the dominant bridge for Caption shifts from the Triple group to the Pair group across both backbones.}
\label{tab:caption_text_control}
\end{table}

\subsection{Out-of-Distribution Validation of Shared-Neuron Saliency}
\label{app:ood_shared_groups}

To examine whether the shared-neuron patterns identified on COCO generalize beyond the in-domain benchmark, we repeat the same group-level masking analysis on multiple out-of-distribution datasets, including GQA for VQA, TextVQA for OCR, and Flickr30k for Caption and Retrieval. Table~\ref{tab:ood_shared_groups} reports the relative performance drop after masking task-specific, best pair-shared, best triple-shared, and general neuron groups. Overall, the results remain qualitatively consistent with the in-domain findings: the most salient shared groups continue to induce larger performance drops, while task-specific and general groups have relatively weaker effects. This suggests that the shared-neuron patterns identified on COCO do not merely depend on a particular data distribution, but remain stable on related benchmarks with distribution shift, further supporting that these routing-and-neuron structures reflect more robust cross-task computation patterns.

\begin{table}[t]
\centering
\footnotesize
\renewcommand{\arraystretch}{1.12}
\setlength{\tabcolsep}{2.0pt}
\begin{tabular}{lcccc}
\toprule
\textbf{Task} & \textbf{Spec.} & \textbf{Pair} & \textbf{Triple} & \textbf{Gen.} \\
\midrule
GQA       &  6.90 & \makecell{36.50\\(VQA+Retrieval)} & 1.20 & 0.62 \\
TextVQA   &  3.40 & \makecell{21.00\\(VQA+OCR)} & 1.50 & 1.05 \\
Flickr30k-Cap & 11.00 & 6.30 & \makecell{24.80\\(VQA+OCR\\+Caption)} & 3.00 \\
Flickr30k-Ret &  1.60 & \makecell{5.35\\(VQA+Retrieval)} & 1.10 & 1.20 \\
\bottomrule
\end{tabular}
\caption{Out-of-distribution validation of shared-neuron saliency. Values denote relative performance drop (\%) after masking different neuron groups, measured by Accuracy for VQA-GQA, ANLS for OCR-TextVQA, BLEU-4 for Caption-Flickr30k, and NDCG@5 for Retrieval-Flickr30k. Spec., Pair, Triple, and Gen. denote task-specific, best pair-shared, best triple-shared, and four-task shared groups, respectively. Larger values indicate stronger causal importance.}
\label{tab:ood_shared_groups}
\vspace{-4mm}
\end{table}

\begin{figure*}[!p]
    \centering
    \captionsetup[subfigure]{justification=centering, skip=3pt}
    \captionsetup{font=small, skip=4pt}

    \begin{subfigure}[t]{0.96\textwidth}
        \centering
        \includegraphics[width=\textwidth]{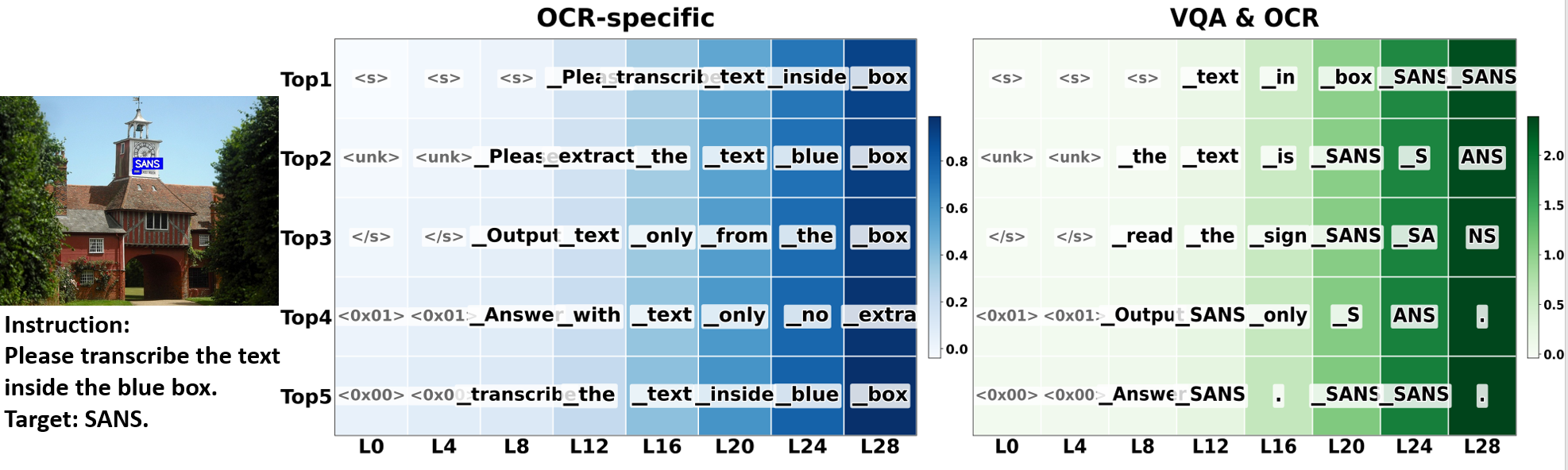}
        \vspace{1.5mm}
        \caption{OCR case study in \textsc{LLaVA}-1.5.}
        \label{fig:appendix_logitlens_ocr}
    \end{subfigure}

    \vspace{0.8mm}

    \begin{subfigure}[t]{0.96\textwidth}
        \centering
        \includegraphics[width=\textwidth]{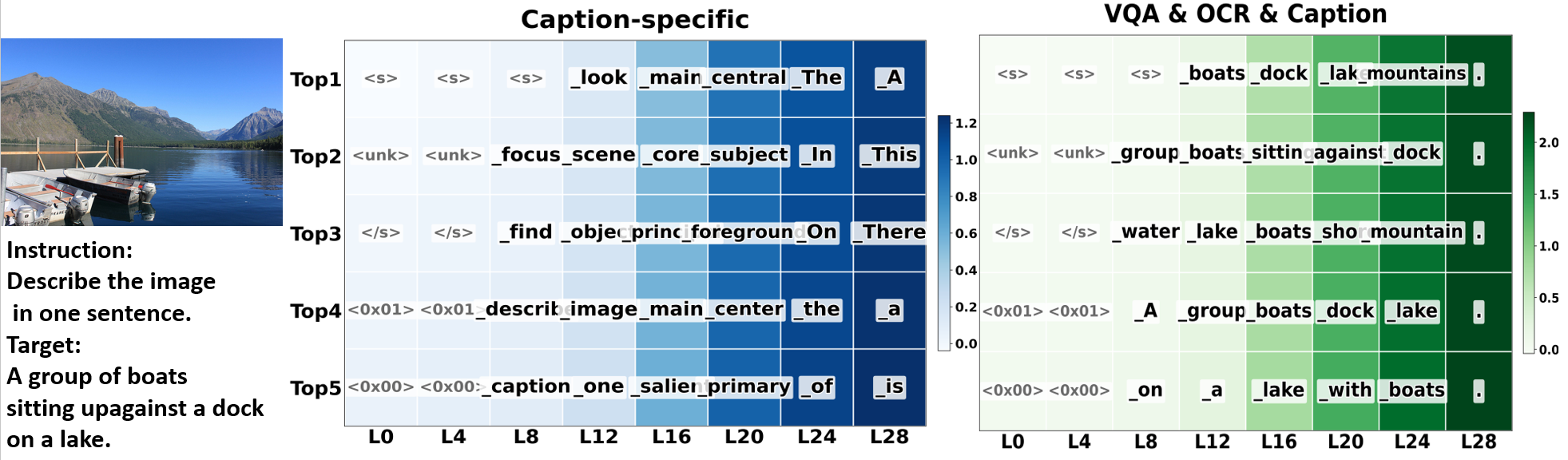}
        \vspace{1.5mm}
        \caption{Caption case study in \textsc{LLaVA}-1.5.}
        \label{fig:appendix_logitlens_caption}
    \end{subfigure}

    \vspace{0.8mm}

    \begin{subfigure}[t]{0.96\textwidth}
        \centering
        \includegraphics[width=\textwidth]{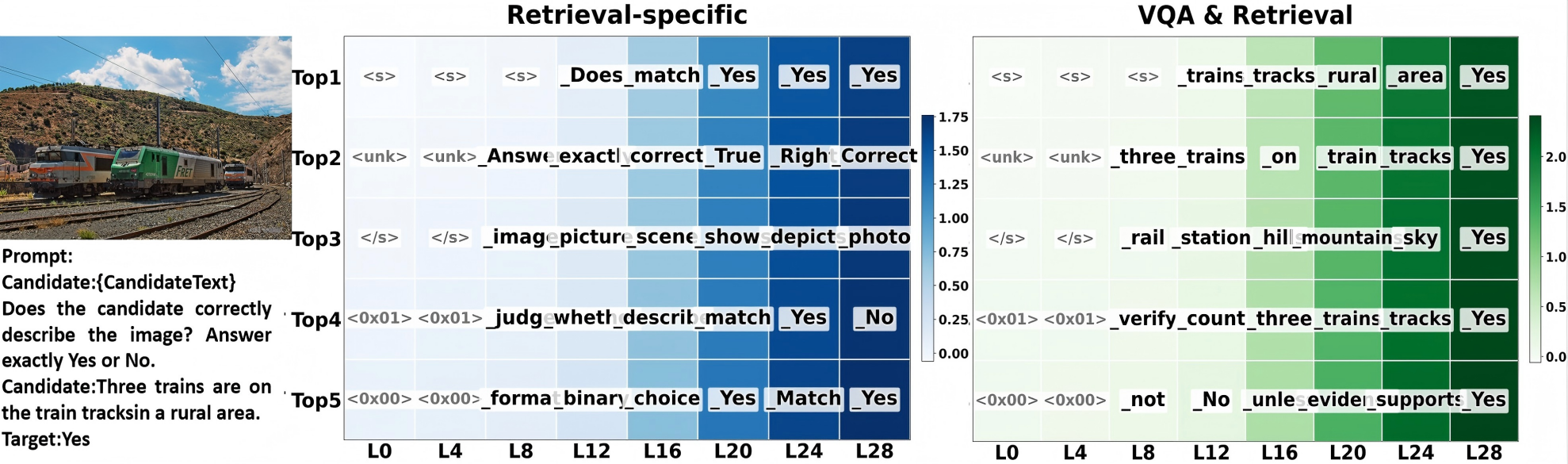}
        \vspace{1.5mm}
        \caption{Retrieval case study in \textsc{LLaVA}-1.5.}
        \label{fig:appendix_logitlens_retrieval}
    \end{subfigure}

    \caption{Logit Lens case studies in \textsc{LLaVA}-1.5.
    Rows show Top-5 tokens and columns are sampled every 4 layers; color indicates $\Delta$logit (baseline$-$masked).
    \textbf{(a)} OCR compares the OCR-specific group and the VQA\&OCR shared group.
    \textbf{(b)} Caption compares the Caption-specific group and the VQA\&OCR\&Caption shared group.
    \textbf{(c)} Retrieval compares the Retrieval-specific group and the VQA\&Retrieval shared group.}
    \label{fig:appendix_logitlens_all}
\end{figure*}

\subsection{Logit Lens Evidence: Shared Bridges as Anchors, Task Neurons as Scaffolds}
\label{sec:appendix_logitlens}

\paragraph{Visualization Setup.}
We perform an MLP-based Logit Lens visualization on single representative cases. At each transformer layer $\ell$, we project the output of the MLP block directly to the vocabulary using the frozen language head to obtain the immediate token distribution.
We compare forward passes under three settings: (i) baseline, (ii) masking the dominant salient neuron group, and (iii) masking the Task-specific group.
We report $\Delta$Logit (baseline $-$ masked), where a positive value indicates the neuron group's contribution to promoting a specific token.
For clarity, we display the Top-5 tokens with the largest $\Delta$Logit increase at representative layers.

\paragraph{OCR}
As shown in Fig.~\ref{fig:appendix_logitlens_ocr}, the OCR-only group acts as a \emph{mode switcher}, boosting meta-tokens related to instruction adherence and output formatting (e.g., \textit{transcribe/text/inside/reads}), thereby setting the operational state for text extraction.
The Shared Bridge (VQA\&OCR), however, performs the critical \textbf{visual-symbol grounding}, directly promoting tokens corresponding to the target string content (e.g., subword pieces of ``SANS''), confirming its role as the character recognition engine.

\paragraph{Caption}
As shown in Fig.~\ref{fig:appendix_logitlens_caption}, the Caption-only group constructs the \emph{structural scaffolding}. It boosts tokens related to \textbf{saliency priors} (e.g., \textit{main/central/core}) to direct attention, and discourse markers (e.g., \textit{The/A/In}) to maintain syntactic fluency.
Conversely, the Shared Bridge (VQA\&OCR\&CAP) functions as the \textbf{semantic filler}, strongly boosting content words that map to specific image entities (e.g., \textit{boats/dock/lake}), populating the syntactic template with visual evidence. In contrast to task-specific neurons that mainly shape format and fluency, this bridge contributes more directly to content grounding along the core pathway.

\paragraph{Retrieval}
As shown in Fig.~\ref{fig:appendix_logitlens_retrieval}, the Retrieval-only group enforces \emph{format adherence}, primarily boosting binary decision tokens (e.g., \textit{Yes/True/Match}) and generic visual terms, imposing a task-level \textbf{alignment bias}.
Crucially, the Shared Bridge (VQA\&RET) acts as the \textbf{semantic verifier}. Before the final layer, it explicitly activates fine-grained concepts present in the image (e.g., \textit{trains/tracks/rural}) to validate the text candidate, providing the necessary \textbf{evidence} to support the final judgment.

The case visualizations across four tasks exhibit a relatively consistent pattern: task-specific neuron groups tend to boost task format, expression scaffolds, or a coarse candidate space;
shared-bridge neuron groups are more likely to concentrate their boosts on fine-grained semantic tokens consistent with image evidence or the target answer.
This pattern aligns with the cross-task ablation results, providing token-level supporting evidence for the conclusion that “task-only units mainly serve auxiliary adaptation, while shared bridges are closer to the core pathway.”

\section{Supplementary Neuron Steering Results}
\label{app:supp_neuron_steering}

\subsection{Learning-based Neuron Scaling and Baselines}
\label{sec:appendix_steer_method}

\paragraph{Steering target (bank) and evaluation split.}
We perform neuron steering on the same unified COCO-based split used for neuron analysis to keep the setting consistent across discovery, ablation, and editing.
We learn neuron-wise scaling factors on the \textbf{2K dev} split and report performance on the \textbf{3K test} split with all backbone weights frozen.
Instead of scaling the entire top-1\% set, we steer a fixed-budget subset drawn from each task's \emph{dominant salient neuron group} (defined in \S\ref{sec:key-findings} and Appendix~\ref{sec:appendix_finding3}), which improves stability and reduces unintended cross-task interference.

\paragraph{Learnable scaling wrapper.}
For each edited layer, we keep the original MLP weights (\texttt{gate\_proj/up\_proj/down\_proj}) frozen and insert a lightweight wrapper on the intermediate FFN activation.
We introduce a neuron-wise learnable vector \texttt{ns\_scale} together with a binary mask \texttt{ns\_mask} indicating which hidden dimensions belong to the chosen bank; only masked dimensions are learnable, while all others are fixed (scaling factor equals 1).
This design yields a parameter-efficient editing interface while preserving the base model architecture.

\paragraph{Training objective (CE + KL regularization).}
We optimize \texttt{ns\_scale} using a supervised task loss on dev data (cross-entropy with respect to\ ground-truth targets) while adding a KL regularizer that constrains the edited model's output distribution to stay close to the unedited backbone (teacher).
The KL term effectively prevents overly aggressive edits and improves the stability of gains across tasks and settings.

\paragraph{Baselines (matched budget).}
We compare against train-free and learnable baselines under the same neuron budget:
(i) \textbf{fixed amplification ($\times$2)} on the same bank;
(ii) \textbf{dev-tuned amplification (grid search)} that selects an amplification factor on dev;
(iii) \textbf{learnable scaling (random-neuron bank)} that learns the same number of scales but on random neurons;
(iv) \textbf{w/o KL (CE only)} that drops the KL regularizer.
(LoRA-style low-rank tuning was also explored under a low-budget setting but did not yield competitive improvements, so we omit it from the main comparisons.)

\begin{table}[!t]
\centering
\footnotesize
\renewcommand{\arraystretch}{1.05}
\setlength{\tabcolsep}{4pt}
\resizebox{\columnwidth}{!}{%
\begin{tabular}{llcccc}
\toprule
\textbf{Task} & \textbf{Baseline} & \textbf{Mean $\Delta$ (pp)} & \textbf{SD (pp)} & \textbf{p-value} \\
\midrule
\multirow{3}{*}{VQA} 
& Base              & 2.15 & 1.00 & *** \\
& Fixed-Amp & 1.25 & 0.96 & *** \\
& RandNeuron       & 2.05 & 1.04 & *** \\
\midrule
\multirow{3}{*}{OCR}
& Base              & 2.59 & 1.10 & *** \\
& Fixed-Amp & 1.99 & 1.20 & *** \\
& RandNeuron       & 2.84 & 1.16 & *** \\
\midrule
\multirow{3}{*}{Caption}
& Base              & 1.22 & 0.90 & *** \\
& Fixed-Amp & 1.17 & 0.92 & *** \\
& RandNeuron       & 1.47 & 0.94 & *** \\
\midrule
\multirow{3}{*}{Retrieval}
& Base              & 2.98 & 0.96 & *** \\
& Fixed-Amp & 2.56 & 1.04 & *** \\
& RandNeuron       & 2.81 & 1.00 & *** \\
\bottomrule
\end{tabular}%
}
\caption{Performance difference $\Delta$ (pp) between our bridge-neuron rescaling method and the baselines (Base, Fixed-Amp, and RandNeuron). Statistical significance is assessed via paired bootstrap resampling ($B{=}1000$, $K{=}500$). Asterisks indicate significance levels; *** denotes $p<0.001$ (extremely significant).}
\label{tab:llava_significance}
\vspace{-6mm}
\end{table}

\subsection{OOD Benchmarks for Generalizability}
\label{sec:appendix_steer_ood}

\paragraph{GQA (OOD for VQA).}
GQA is a visual reasoning benchmark built to reduce superficial biases and emphasize compositional reasoning, with questions derived from scene-graph structure and controlled answer distributions \citep{Hudson_2019_CVPR}.
We use it to test whether the learned scaling remains effective beyond the COCO-based evaluation distribution for VQA-style judgments.

\paragraph{TextVQA (OOD for OCR/text understanding).}
TextVQA focuses on questions that require reading and reasoning over scene text, pairing VQA-style prompts with OCR-relevant visual evidence \citep{Singh_2019_CVPR}.
We use it as an out-of-distribution testbed for our OCR-related steering behavior.

\paragraph{Flickr30k (OOD for Caption and Retrieval).}
Flickr30k contains images with multiple human-written captions and is widely used for captioning and retrieval \citep{young-etal-2014-image}. We use it to test the OOD generalization of our Caption/Retrieval steering under distribution shift.

\paragraph{OOD tuning protocol.}
For each OOD benchmark, we keep the neuron bank fixed (i.e., we do not re-localize heads or re-rank neurons) and only re-learn scaling factors on a small \textbf{20\%} subset of the OOD data, then evaluate on the remaining \textbf{80\%}.
We additionally compare against (i) train-free amplification and (ii) directly transferring scales learned on the COCO-based dev split, to isolate the benefit of lightweight in-domain scale learning in OOD settings.

\subsection{Statistical Significance Tests}
\label{sec:appendix_significance}

We conduct statistical significance tests for the \emph{bridge-neuron steering} results reported in Table~\ref{tab:steer_effectiveness}. For each task, we compare Ours against three baselines: Base, Fixed-Amp ($\times$2; a train-free uniform amplification baseline), and RandNeuron (learned scaling on a random neuron set matched in size to our bridge-neuron set). Let $\mathcal{D}=\{x_i\}_{i=1}^{N}$ denote the test instances of a task (image-level instances for VQA/OCR/Caption; query-level instances for Retrieval). We perform paired bootstrap resampling with $B{=}1000$ resamples. For each resample $b$, to reduce computation and ensure a consistent bootstrap scale across tasks, we sample $K{=}500$ indices with replacement from $\{1,\dots,N\}$ to form a multiset $\mathcal{D}^{(b)}$, and evaluate \textsc{Ours} and each baseline on the same $\mathcal{D}^{(b)}$ to preserve pairing. For each baseline $j$, we define the resample-level improvement as the difference between the task metric of \textsc{Ours} and that of $j$ on $\mathcal{D}^{(b)}$, where the metric is Acc.\ for VQA, ANLS for OCR, BLEU-4 for Caption, and NDCG@5 for Retrieval; for Caption, we compute corpus-level BLEU-4 on the resampled subset. We report the mean and sample standard deviation over the $B$ resamples as the ``Mean $\Delta$'' and ``SD'' columns, and compute a two-sided bootstrap p-value by doubling the smaller of the two empirical tail probabilities.

As shown in Table~\ref{tab:llava_significance}, our bridge-neuron rescaling yields statistically significant improvements over all three baselines across all four tasks on LLaVA-1.5 7B. Paired bootstrap tests further confirm the robustness of these gains, with all p-values being extremely small ($p<0.001$), indicating strong statistical significance.

\section{Computational Resources.}
All experiments were conducted on a single machine equipped with two NVIDIA Tesla V100 GPUs (32GB memory each), an Intel(R) Xeon(R) Gold 6230R CPU running at 2.10GHz (8 cores), and 64GB of RAM.
Our interpretability pipeline—including head-level targeted causal ablations for routing localization, neuron attribution, cross-task/group-level neuron masking, and lightweight learnable neuron scaling for steering—was implemented in PyTorch with Hugging Face Transformers.

\begin{table*}[t]
\centering
\renewcommand{\arraystretch}{1.2}
\setlength{\tabcolsep}{6pt}
\resizebox{\textwidth}{!}{%
\begin{tabular}{l|l|l}
\toprule
\textbf{Task} & \textbf{Model} & \textbf{Prompt / Input Format} \\
\midrule
\multirow{2}{*}{VQA}
& LLaVA-1.5 &
\makecell[l]{\textbf{USER:}\\
\texttt{<image>}\\
\texttt{Question: \{Question\}}\\
\texttt{Answer the question briefly using one or a few words.}\\
\textbf{ASSISTANT:}} \\
\cmidrule(lr{0.5em}){2-3}
& Qwen2.5-VL &
\makecell[l]{\textbf{Structured chat message (via }\texttt{processor.apply\_chat\_template}\textbf{):}\\
\texttt{[ \{role: user, content: [}\\
\texttt{\quad \{type: image, image: \{Image\}\},}\\
\texttt{\quad \{type: text, text: ...\}}\\
\texttt{] \} ]}\\
\textbf{Text:}\\
\texttt{Question: \{Question\}}\\
\texttt{Answer the question with a short phrase.}} \\
\midrule

\multirow{2}{*}{OCR}
& LLaVA-1.5 &
\makecell[l]{\textbf{USER:}\\
\texttt{<image>}\\
\texttt{Please transcribe the text inside the blue box.}\\
\texttt{Output text only.}\\
\textbf{ASSISTANT:}} \\
\cmidrule(lr{0.5em}){2-3}
& Qwen2.5-VL &
\makecell[l]{\textbf{Structured chat message (via }\texttt{processor.apply\_chat\_template}\textbf{):}\\
\texttt{[ \{role: user, content: [}\\
\texttt{\quad \{type: image, image: \{Image\}\},}\\
\texttt{\quad \{type: text, text: ...\}}\\
\texttt{] \} ]}\\
\textbf{Text:}\\
\texttt{Transcribe the text inside the blue box.}\\
\texttt{Output text only.}} \\
\midrule

\multirow{2}{*}{Caption}
& LLaVA-1.5 &
\makecell[l]{\textbf{USER:}\\
\texttt{<image>}\\
\texttt{Describe the image in one sentence.}\\
\textbf{ASSISTANT:}} \\
\cmidrule(lr{0.5em}){2-3}
& Qwen2.5-VL &
\makecell[l]{\textbf{Structured chat message (via }\texttt{processor.apply\_chat\_template}\textbf{):}\\
\texttt{[ \{role: user, content: [}\\
\texttt{\quad \{type: image, image: \{Image\}\},}\\
\texttt{\quad \{type: text, text: ...\}}\\
\texttt{] \} ]}\\
\textbf{Text:}\\
\texttt{Describe the image in one sentence.}} \\
\midrule

\multirow{2}{*}{Retrieval (Pairwise Matching)}
& LLaVA-1.5 &
\makecell[l]{\textbf{USER:}\\
\texttt{<image>}\\
\texttt{Candidate: \{CandidateText\}}\\
\texttt{Does the candidate correctly describe the image?}\\
\texttt{Answer exactly Yes or No.}\\
\textbf{ASSISTANT:}} \\
\cmidrule(lr{0.5em}){2-3}
& Qwen2.5-VL &
\makecell[l]{\textbf{Structured chat message (via }\texttt{processor.apply\_chat\_template}\textbf{):}\\
\texttt{[ \{role: user, content: [}\\
\texttt{\quad \{type: image, image: \{Image\}\},}\\
\texttt{\quad \{type: text, text: ...\}}\\
\texttt{] \} ]}\\
\textbf{Text:}\\
\texttt{Candidate: \{CandidateText\}}\\
\texttt{Does the candidate correctly describe the image?}\\
\texttt{Answer exactly Yes or No.}} \\
\bottomrule
\end{tabular}%
} 
\vspace{-6pt}
\caption{Prompt and input formats used for evaluation. LLaVA-1.5 is shown in its common textual chat style, while Qwen2.5-VL uses structured multimodal chat messages serialized by the processor chat template. For retrieval, candidates are ranked by the model probability of the \texttt{Yes} response (pairwise matching).}
\label{tab:prompt_templates_eval}
\end{table*}

\end{document}